\documentclass[twoside]{article}

\usepackage{PRIMEarxiv}			
\renewcommand{\normalsize}{\fontsize{12}{14}\selectfont} 
\usepackage{adjustbox} 			
\usepackage[dutch, american]{babel}	
\usepackage{caption}				
\usepackage{enumerate}			
\usepackage{mathtools}			
\usepackage{pdflscape}				
\usepackage{xltabular}				
\usepackage{tabularx}				
\usepackage{graphicx}				
\usepackage{url}					
\usepackage{authblk}				

\usepackage{hyperref}				
\hypersetup{
    hidelinks 						
}

\usepackage{doi}

\usepackage[natbibapa]{apacite}		

\newcommand\blfootnote[1]{
    \begingroup
    \renewcommand\thefootnote{}\footnote{#1}
    \addtocounter{footnote}{-1}
    \endgroup
}

\hypersetup{
    pdftitle={Unsupervised Acquisition of Discrete Grammatical Categories},
    pdfauthor={David Ph. Shakouri, Crit Cremers, Niels O. Schiller},
    pdfkeywords={multiagent systems, modelling language acquisition, knowledge acquisition, statistical natural language processing, rule learning, language acquisition, graph theory for NLP, machine learning, logic-based artificial intelligence (AI), language models, syntactic categories, agglomerative hierarchical clustering (AHC), cluster analysis, clustering}
}

\title{Unsupervised Acquisition of Discrete Grammatical Categories}

\author[1,2]{\textbf{David Ph. Shakouri}}
\author[1]{\textbf{Crit Cremers}}
\author[1,2,3]{\textbf{Niels O. Schiller}}

\affil[1]{Leiden University Centre for Linguistics (LUCL), Leiden University}
\affil[2]{Leiden Institute for Brain and Cognition (LIBC), Leiden University}
\affil[3]{City University of Hong Kong (CityU), Hong Kong}

\date{}						

\begin{document}
\maketitle\blfootnote{Unsupervised Acquisition of Discrete Grammatical Categories © 2025 by David Ph. Shakouri, Crit Cremers, Niels O. Schiller is licensed under CC BY-NC-ND 4.0}

\begin{abstract}
This article presents experiments performed using a computational laboratory environment for language acquisition experiments. It implements a multi-agent system consisting of two agents: an adult language model and a daughter language model that aims to learn the mother language. Crucially, the daughter agent does not have access to the internal knowledge of the mother language model but only to the language exemplars the mother agent generates. These experiments illustrate how this system can be used to acquire abstract grammatical knowledge. We demonstrate how statistical analyses of patterns in the input data corresponding to grammatical categories yield discrete grammatical rules. These rules are subsequently added to the grammatical knowledge of the daughter language model. To this end, hierarchical agglomerative cluster analysis was applied to the utterances consecutively generated by the mother language model. It is argued that this procedure can be used to acquire structures resembling grammatical categories proposed by linguists for natural languages. Thus, it is established that non-trivial grammatical knowledge has been acquired. Moreover, the parameter configuration of this computational laboratory environment determined using training data generated by the mother language model is validated in a second experiment with a test set similarly resulting in the acquisition of non-trivial categories.
\end{abstract}

\section{Introduction}
\label{introduction}

This article presents a case study on the acquisition of discrete grammatical categories, which has been executed using a multi-agent computational laboratory environment to simulate language acquisition experiments. This system has been named \textsc{modoma}, an acronym for \textit{moeder-dochter-machine} (Dutch for ‘mother-daughter-machine’). The \textsc{modoma} implements language acquisition as the result of an interaction between an adult language model, the mother, and a daughter language model, which is designed to acquire grammatical knowledge of the target language. The mother agent is based on the \textsc{delilah} language model generating and parsing Dutch utterances (\citeauthor{Cremers:1995}, \citeyear{Cremers:1995}, cf. \citeauthor{Cremers:2014}, \citeyear{Cremers:2014}). During the interaction the mother agent presents sample sentences of the target language to the daughter agent. As a result, the daughter language model is defined and develops the abilitity to produce and analyze utterances corresponding to her developing grammar. Similarly to human children acquiring their mother languages, the daughter language model is able to process utterances before it has fully acquired a grammar describing the target language.

Building on the work by \citet{Shakouri:2025}, the goal of this study is to model the acquisition of discrete grammatical categories corresponding to parts of speech such as noun, verb or adjective. The first objective is to assess whether a statistical data analysis technique, established in the literature with respect to language produced by humans, can be applied to \textsc{delilah}’s output. In particular, words can be grouped based on the observation that they occur in similar contexts, see the study by \citet{Schutze:1995} for an early example illustrating this type of approach. Considering the rapid increase in the adoption of language models, examining whether machine-generated language reflects patterns similar to those found in human language is becoming increasingly relevant. The grammatical knowledge resulting from these language acquisition simulations should be represented in such a way that it can be used by the daughter language model to generate and parse utterances. One of the key contributions of this study is that the results of this analysis are used to acquire discrete rule-based representations of grammatical knowledge encoded by graph-based feature-value pairs. In this framework, each linguistic feature such as the classification of syntactic categories is represented as a node in the graph, while the relationships between these features and their corresponding values are captured by edges. These graph structures can be used productively by the daughter language model.

In the machine learning literature, a common distinction is made between supervised and unsupervised learning. Supervised learning involves presenting the algorithm with annotated exemplars, whereas unsupervised learning processes raw unannotated data samples \citep[cf.][pp.~16--17]{Duda:2001}. Examples of supervised learning approaches to natural language processing include Data-Oriented Parsing \citep[DOP,][]{Bod:2003} and Memory-Based Language Processing \citep[MBLP,][]{Daelemans:2005}. Crucially, the \textsc{modoma} only employs unsupervised learning: Children acquiring their first language do not have direct access to the grammatical knowledge of the adults providing examples of the target language. Hence, application of unsupervised learning is in accordance with the design requirements of the \textsc{modoma}. Similarly, \textsc{delilah} \citep{Cremers:1995}, the mother in the mother-daughter system, produces sentences, which serve as input to the language acquisition system of the daughter agent and although the mother language model employs a grammar, which assigns parses and labels to the constituents represented by the sentences she produces, these labels are not provided to the daughter. The experiments described in this article utilize clustering to model the acquisition representations of grammatical categories, namely parts of speech. In this respect, one advantage of clustering over supervised techniques such as classification is that it provides an unsupervised approach for acquiring linguistic knowledge.

Interestingly, as soon as the daughter agent has learned labels specifying her grammar, she can use this acquired knowledge to annotate previously seen and new utterances by the mother agent. In the context of the \textsc{modoma}, this acquisition technique has been called internal annotation. Internal annotation implements a form of self-supervised learning (cf. {\citeauthor{Balestriero:2023}, \citeyear{Balestriero:2023}, \citeauthor{Gui:2024}, \citeyear{Gui:2024} for surveys and \citeauthor{Brown:2020}, \citeyear{Brown:2020}, \citeauthor{Devlin:2019}, \citeyear{Devlin:2019}, \citeauthor{Lan:2020}, \citeyear{Lan:2020}, \citeauthor{Orhan:2020}, \citeyear{Orhan:2020}, \citeauthor{Yarowsky:1995}, \citeyear{Yarowsky:1995} for some influential and notable examples). In particular, the daughter agent can employ previously acquired linguistic knowledge to acquire more complex structures. The difference between supervised learning and internal annotation is that in the case of supervised acquisition techniques the labels are provided externally with respect to the daughter, for instance by another language model such as the mother, a trained linguist, or an annotated corpus. Conversely, for internal annotation the labels are provided internally, that is, based on previously acquired grammatical knowledge by the daughter. Thus, if the annotations can be provided by the daughter, during later stages of acquisition the daughter can employ supervised acquisition techniques as well. Nonetheless, notwithstanding the use of supervised techniques the entire acquisition system is unsupervised. 

\section{Description of the \textsc{Modoma}}
\label{description}

The \textsc{modoma} system is the result of a research project aimed at providing a laboratory approach to language acquisition. This project builds on the \textsc{delilah} project, which resulted in the Leiden parser and generator of Dutch, a language model generating and parsing Dutch utterances such as sentences or noun phrases (NPs). The \textsc{modoma} consists of three main components: (1) \textsc{delilah} providing samples of the adult language, (2) a daughter language model, which is developed to acquire the mother language and starts communicating as soon as it has acquired any suitable construction, and (3) an interaction system enabling the conversation, that is, the exchanging of utterances between the mother and the daughter agents \citep[cf.][]{Shakouri:2025}. These three components will be briefly discussed.

\textsc{Delilah} implements a combinatory list grammar, a formalism related to combinatory categorial grammar (CCG, \citeauthor{Cremers:2002}, \citeyear{Cremers:2002}, pp.~378-386; \citeauthor{Cremers:2014}, \citeyear{Cremers:2014}, pp.~115-137, cf. \citeauthor{Baldridge:2003}, \citeyear{Baldridge:2003} for surveys \citeauthor{Moortgat:1997}, \citeyear{Moortgat:1997}, \citeauthor{Steedman:1996}, \citeyear{Steedman:1996}). \citet{Cremers:2014} present detailed discussions and analyses of the design and formalisms employed by \textsc{delilah}, see also \citet[pp.~25--29]{Reckman:2009}\footnote{A demonstration version presenting functionalities of this complete language model is provided at https://delilah.universiteitleiden.nl/indexen.html \citep{Cremers:2025}, last accessed Jan 15, 2025.}. \textsc{Delilah}'s generator and parser execute a grammar containing graph structures representing lexical items such as Dutch words and constructions. These graphs explicitly represent grammatical properties of these words and constructions such as the phonological form, concepts, logical forms encoding meaning \citep[cf.][]{Cremers:2008}, grammatical number, grammatical person, and the syntactic category (e.g., determiner, noun, or verb). This predefined grammar model is used to produce composed structures corresponding to sentences or constituents (e.g., noun phrases or verb phrases) and analyze utterances by unifying these graph structures. This type of approach to generating and parsing language is similar to the approach by Head-Driven Phrase Structure Grammar \citep[HPSG,][]{Pollard:1987,Pollard:1994}. \textsc{Delilah}’s generator takes as its input a word or concept representation and produces a utterance based on this input while for \textsc{delilah}’s parser the input is a surface form string and the output is a parse providing a structural analysis of this input string. In both cases the output is represented by a template corresponding to a well-formed structure according to \textsc{delilah}’s grammar. When \textsc{delilah} is called by the \textsc{modoma}, this template corresponds to a utterance, which is used to take part in the conversation with the daughter agent. For the purposes of the \textsc{modoma} project, \textsc{delilah} has been taken as is: This language model provides the samples of the adult language while the project is aimed at implementing a language acquisition system.

The daughter agent is a novel language model that has been specifically designed and developed for this project. It consists of three main components: (1) 
a generator and a parser, which enable the daughter agent to generate utterances and provide analyses as well as grammaticality judgments with respect to utterances generated by the mother language model, (2) a grammar, which is used by the generator and parser to process language and consists of templates corresponding to acquired words and grammatical constructions, and (3) a language acquisition device enabling the acquisition of knowledge of the target language and subsequently adding this newly acquired knowledge to the grammar. These components are connected to facilitate acquiring knowledge of the target language, representing it in the grammar, and using this grammar model to process utterances. Thus, language acquisition results in the construction of a functioning language model.

\begin{figure*}[t]																
	\centering
	\[																	
		\begin{bmatrix*}[l]													
			\textsc{memory stack position:} & \text{0}\\								
			\textsc{lexical entry number:} & \text{1}\\
			\textsc{session id:} & \text{1601581107338}\\
			\textsc{confidence lexical entry:} & \text{60}\\
			\textsc{head directionality:} & \text{null}\\
			\textsc{confidence head directionality:} & \text{0}\\
			\textsc{terminal:} & \text{T}\\
			\textsc{phonform:} & \textit{winkelen}\\
			\textsc{semform:} & \text{A}\\
			\textsc{semform index:} & \text{null}\\
			\textsc{grammatical properties:} &
			\begin{bmatrix*}[l]												
				\textsc{property type:} & \textsc{\textsc{a}}\\							
				\textsc{property value:} & \text{n}\\
				\textsc{confidence property:} & \text{60}\\
			\end{bmatrix*}\\													
				\textsc{head:} & \begin{bmatrix*}[l] \textsc{winkelen.1} \end{bmatrix*}\\			
				\textsc{argument:} & \begin{bmatrix*}[l] \text{null} \end{bmatrix*}\\			
		\end{bmatrix*}														
	\]																	
	\caption{Example of the representation of an acquired lexical item for \textit{winkelen} (‘shop’, \textsc{v}) by the daughter agent}	
	\label{fig:item_daughter}												
\end{figure*}		

Similarly to \textsc{delilah}, the templates in the grammar are represented by graph structures, which explicitly specify the grammatical properties of the items. Figure \ref{fig:item_daughter} presents an example of a minimal template acquired by the daughter language model. These templates consist of binary graph structures containing a \textsc{head} and an \textsc{argument} position, which are designed to hold graphs of the same type as the main graph structure. Moreover, grammatical properties of the items are explicitly represented using feature-value pairs. For example, these graph structures include a phonological representation (\textsc{phonform}), a semantic representation (\textsc{semform}) consisting of a unique label representing the item's denotation, an indication of whether the item is a word or a construction identified by the \textsc{terminal} feature, the \textsc{head directionality} (i.e., currently unknown, right-headed, or left-headed) used to linearize the structure in order to generate an utterance, and the \textsc{semform index}, which allows implementing shared referents between items to model (the acquisition of) linguistic phenomena such as reflexives and anaphora. In addition, the confidence the language model has with respect to previously acquired items and grammatical properties is represented by confidence feature-value pairs such as \textsc{confidence lexical entry} (for the whole item) and \textsc{confidence head directionality} (for the head directionality). Crucially, these templates enable the specification of a dynamic list of feature-value pairs to represent the results of language acquisition procedures. This list can be employed to model the acquisition of a wide range of linguistic phenomena \citep[cf.][for experiments using this list to encode functional and content categories]{Shakouri:2025}. As the \textsc{modoma} implements unsupervised learning, the daughter language model has no information regarding the labels used by the mother language model. Therefore, the results of language acquisition are represented by unique alpha-numeric labels such as $\langle\mbox{\textsc{a}:n}\rangle$ (for feature \textsc{a} and value \textit{n}) for instance corresponding to $\langle\mbox{\textsc{partofspeech}:verb}\rangle$.\footnote{Technically, the features in the dynamic list of feature-value pairs are implemented as values of \textsc{property type}. However, functionally they represent features with the corresponding value of \textsc{property value} acting as their value.} In particular, the linguistic knowledge resulting from the language acquisition experiments presented by this article is saved using this list. Finally, technical metadata are similarly retained for example the combination of the \textsc{session id} and \textsc{lexical entry number} can be used to uniquely identify each item of acquired knowledge by the \textsc{modoma}.

Typically computational approaches to natural language processing using graph structures to represent linguistic knowledge do not make a principled distinction between the lexicon and the grammar. In more conventional grammar descriptions, the term lexicon usually refers to the acquired and retained word list and the properties of the contained words such as grammatical features, phonological form and reference (meaning) while the term grammar is reserved for the collection of abstract rules describing the language. In computational approaches employing graph structures words are usually saved using the same type of templates as rules while for rules these templates display a higher level of abstraction, that is, less or no specific words with for instance a fully defined phonological form are specified in the saved graph. Thus, for the algorithm there is no difference between what is traditionally referred to as lexicon versus grammar. For this reason, the terms grammar and lexicon are used as synonyms for the linguistic knowledge. In this respect, the \textsc{modoma} daughter language model resembles HPSG \citep{Pollard:1987,Pollard:1994} and the \textsc{delilah} system, which is used as the mother agent: The result of language acquisition is either a template corresponding to a word or a specification of a feature-value pair restricting the combinatory properties of a new or previously acquired template \citep[cf.][]{Shakouri:2025}.

The generator and parser of the daughter language model generate and analyze utterances by searching the grammar containing templates previously constructed by the language acquisition device and subsequently unifying graphs to create well-formed structures in accordance with the currently acquired grammar model. Similar to human children, the language learning agent should be able to employ a grammar and language it has not fully acquired yet. Therefore, from the perspective of simulating first language acquisition an important characteristic is the \textsc{modoma} daughter language model’s ability to process unknown words and constructions. Crucially, the grammatical properties represented by these templates including the feature-value pairs in the dynamic list impose restrictions on which templates the generator or parser can combine, thereby limiting the set of well-formed parses and utterances of the target language. Thus, the knowledge representations enable a comprehensive language model capable of producing and analyzing utterances of the target language. Moreover, employing a language model with explicitly defined knowledge allows for the inspection of its contents. Thus, the results and development of language acquisition can be directly evaluated. Therefore, the language model is inherently an accountable artificial intelligence system, ensuring transparency and responsibility in its processes. Employing these types of models provides an interesting perspective in relation to for instance large language models.

Although \textsc{delilah} is a language model employing graph structures to represent grammatical knowledge as well, the daughter agent has no direct information on the mother agent’s analyses other than what she can determine by analyzing \textsc{delilah}’s output utterances. Therefore, defining abstract rules explaining noisy linguistic input data is a computationally complicated task. To this end, the language acquisition device employs a hybrid approach to language acquisition employing statistical as well as rule-based techniques to specify discrete and explicitly defined graph structures. Besides adding new items to the grammar such as learning a new word or construction, language acquisition should result in encoding grammatical properties of previously acquired items. Thus, language acquisition simulations by this system typically define a grammar model that is more specific than the previous version. In particular, these newly added grammatical properties restrict the possible combinations of templates during generation and parsing: As there is more knowlege of the target language, less utterances are considered well-formed. While previous studies demonstrated that functional and content categories can be acquired by a daughter language model \citep[cf.][]{Shakouri:2025}, the research presented in this article focuses on the hybrid acquisition of discrete categories encoding parts of speech, which can be used to subcategorize the items in the grammar. 

Summarizing, while the grammar is specified as a result of language acquisition, the properties of the parser and generator are not changed during language acquisition. Thus, language acquisition results in two types of output (1) a lexicon/grammar and (2) a set of well-formed utterances allowed by this grammar. In this respect, the structure of the daughter language model reflects a basic computational model of natural language processing. According to this model, natural language processing involves three elements:
\begin{enumerate}[(1)]
\item a language: the (indefinite) set of all utterances in the target language;
\item the grammar: a model describing this indefinite set of utterances;
\item an automaton: the parser and generator executing the grammar to produce and evaluate (parse) the utterances corresponding to the language.
\end{enumerate}
The language model that processes the set of utterances, consists of a combination of the automaton and the grammar. Conversely, as a result of language acquisition only the grammar model is defined and updated by the language acquisition device. The present study discusses experiments simulating the acquisition of non-trivial grammatical knowledge modelling parts of speech and utilizing this knowledge to define the grammar model.

Finally, the interaction system manages the exchange of utterances in the multi-agent system, for instance by taking care of turn taking and implementing optional feedback. This novel system enables the investigation of a wide range of theories and the execution of a diverse array of experiments. By focusing on the acquisition of discrete categories, this study aims to demonstrate the viability of the concept and lays the foundation for future investigations into the multi-agent modelling of first language acquisition. The goal is to implement a system, which can be used by researchers to test their models or assumptions regarding language acquisition by executing computational simulations. This article presents the results of such an experiment. An important advantage of providing a computational laboratory simulation of language acquisition is that this enables experiments that would be impossible or unethical to perform using human subjects. Crucially, all components of language acquisition including both agents are part of the same system. Thus, all aspects of the system can be fully configured depending on the experiment using parameter settings and all input, output, exchange of utterances, and processing by the system are logged and can be retrieved by researchers.

As the \textsc{modoma} system provides a computational laboratory simulation of language acquisition, many settings of the system are parametrized and user-controlled. Thus, for each experiment conducted using this system an optimal set of parameter settings has to be determined by the researchers. An example of a parameter is the number of mother utterances the daughter agent should have processed before executing the acquisition of grammatical categories. Crucially, the selected parameter settings do not imply that this specific configuration plays a part in the process of language acquisition, but rather these settings are implemented to enable simulations to assess the role of phenomena proposed to be a factor in language acquisition. Thus, the \textsc{modoma} can contribute to ongoing scientific debates on the role of properties of the input data and language acquisition process by providing a parametrized system.

Moreover, as the \textsc{modoma} constitutes an experimentation environment, all aspects of the system are recorded and can be retrieved. Examples of data that are logged, are: the parameter settings used to configure an experiment, the employed learning techniques, the utterances generated by the agents, and the acquired grammar. Thus, all factors that influence language acquisition such as the generation of samples of the target language by the mother agent, the grammar acquired by the daughter agent, and the acquisition process, are components in a single system and there are no factors that are beyond the control of the researchers. In this respect the \textsc{modoma} laboratory environment shares similarities with those used in the natural sciences as it aims to minimize factors that cannot be controlled or retrieved by the researcher. 
This case study builds on experiments previously conducted as part of the \textsc{modoma} project \citep[cf.][]{Shakouri:2025} and serves as a showcase for the kind of research that can be performed using this laboratory environment. In particular, it resulted in the acquisition of non-trivial grammatical knowledge, which is represented by an explicit grammar model. This defines a language model that can be used to generate and parse sentences.

\section{Related Work}
\label{related_work}

To our knowledge, the combination of design properties of the \textsc{modoma}, that is, two agents in a single computational system such that one agent provides samples of a natural language while the other agent is designed to learn the target language in an unsupervised manner, is unique compared to other approaches in the field of modelling language acquisition. Crucially, this architecture provides novel research possibilities resulting in a computational language acquisition laboratory as researchers can control and measure all aspects of the simulations such as the parameter settings, input to the language acquisition procedures, executed language acquisition processes, produced utterances, and acquired linguistic knowledge by the daughter language model. This section reviews relevant literature and positions our approach within existing work, emphasizing its distinct contributions.

\citet{Hoeve:2022} discuss a road map to model language acquisition employing a teacher-student-loop and carried out two initial experiments related to the first steps. The first experiment focuses on modelling distinct domains using two completely separate vocabularies in an artificial language. The second experiment examines distinctive structures modeled through token repetitions. In these experiments, the teacher selects a fixed amount of training data from a collection of pre-constructed sentences and presents them to multiple students, which are composed of language models. Subsequently, the student language models are tested on a set of other sentences from the artificial language. Moreover, a teacher language model is trained using a different subset of sentences of the artificial language and evaluated. The existing implementation of the framework contrasts with the \textsc{modoma} system particularly with respect to the training and test data, which in the context of the work by \citet{Hoeve:2022} contain preconstructed sentences from an artificial language. Conversely, the \textsc{modoma} employs utterances that are representative of a natural language and are generated online by Delilah using her explicitly defined language model of Dutch. 

The \textsc{modoma} models language acquisition from an ontogenetic perspective, that is, the development of an individual over time. From a phylogenetic perspective, research has been conducted on emergent communication utilizing multi-agent systems. An important example is the Talking heads project by \citet{Steels:2000,Steels:2015}. The Talking heads project is primarily aimed at providing a computational model of the evolutionary origins and development of language. This model employs multiple agents, which take part in language games \citep[cf.][]{Steels:2001}. During a language game, two agents engage in a dialogue typically making reference to their surroundings and as a result of this interaction a language is developed. An example of such a language game is the naming game \citep[cf.][]{Steels:2012}, during which words referring to objects are coined and can spread to the linguistic knowledge of other agents not involved in the initial language game that gave rise to this term. Some crucial differences when compared to the \textsc{modoma} project is that both agents acquire the target language. This contrasts with the \textsc{modoma} experimental design, in which one agent is the adult not acquiring any novel linguistic knowledge and only one agent is updating her linguistic knowledge. Moreover, most experiments conducted within the Talking heads framework employ an artificial language created by the agents themselves as a result of the language games while the adult agent in the \textsc{modoma} experimental set-up outputs utterances based on a grammar model of a natural language, Dutch. Interestingly, while initial studies employing the Talking heads framework focused on the evolutionary emergence of words and lexical development \citep[e.g.,][]{SteelsKaplan:2001}, more recent investigations also addressed syntactic acquisition and development \citep[cf.][]{Steels:2017,Steels:2018,Trijp:2016}.

Moreover, recent work by \citet{Chaabouni:2020} investigates the emergence of artificial language by two deep neural agents involving different architectures such as two gated recurrent units (GRU) and a feed forward network (FFN). In particular, learning speed and generalization accuracy with respect to compositionality are assessed. A Receiver agent receives an input \textit{i} consisting of \textit{$i_{att}$} attributes and \textit{$i_{val}$} possible values and constructs a message \textit{m} based on this input. Subsequently, a Sender agent is inputted with message \textit{m} and produces an output \textit{\^{i}}. This game has met its goals if the input of the Receiver resembles the output of the Sender, that is, \textit{i = \^{i}}. A follow-up study by \citet{Chaabouni:2022} examines the effects of scaling up the ‘dataset, task complexity, and population size’. The study employs a Speaker, which takes an image as its input to generate a message, and a Listener, which receives this message along with a collection of images. The Listener's task is to select the image that corresponds to the message. Interestingly, \citet{Chaabouni:2021} conduct experiments simulating the emergence of discrete color-naming systems. In addition, \citet{Chaabouni:2019} and \citet{Rita:2020} explored the emergence of Zipf’ law with respect to artificial language created by two interacting agents. Similar to the \textsc{modoma} approach, their experimental design involves two interacting agents. Conversely, an important difference is that the language produced by the Receiver and Sender agents consists of ‘simple emergent codes’ while the \textsc{modoma} employs (a fragment of) a natural language. Another significant distinction is the focus on language emergence while the \textsc{modoma} aims at modelling first language acquisition. In this respect these approaches complement each other.

Furthermore, \citet{Griffith:2007} model language evolution using (a population of) Bayesian agents engaged in iterated learning with artificial languages composed of utterance-meaning pairs. Other related research is the Baby SRL project by \citet{Conner:2008,Conner:2009}, which provides computational models for the acquisition of semantic role labeling. There are several notable differences between this line of research and the \textsc{modoma}. For example, in the Baby SRL project samples of the target language are not generated by a language model but are instead sourced from the CHILDES corpus \citep{MacWhinney:2014} or consist of constructed sentences. Moreover, as this model is supervised, the language acquiring algorithms are provided with annotated information on the structure of the target language in contrast to the \textsc{modoma}. 

Alishahi conducted computational experiments with algorithms applied to corpora that have been either artificially generated or produced by humans, to acquire grammatical phenomena. For example, the experiments by \citet{Alishahi:2008} simulate the acquisition of early argument structure, the studies by \citet{Alishahi:2009} focus on lexical category acquisition, and the research by \citet{Alishahi:2012} investigates the acquisition of lexical categories and word meaning. Interestingly, \citet{Matusevych:2013} present methods to improve artificially generated input data representing adult-child interactions, which are intended for use similarly to data from corpora such as CHILDES that include parent-child interactions. However, unlike the \textsc{modoma} experimental design, which aims to simulate all components involved in language acquisition within a system, the algorithm generating the input data is separate from the procedures executing language acquisition. Moreover, \citet{Beekhuizen:2014} present an article on the acquisition of grammatical constructions. Amongst others, this model differs in two interesting ways from the present study: (1) Their algorithm acquires these constructions based on representations of situations and utterances, which have been automatically generated taking into account the distributions in a corpus containing speech directed to children, but (2) unlike the \textsc{modoma} this model does not contain multiple agents.

Finally, \citet{McCoy:2020} have carried out interesting research on the computational modelling of natural language acquisition employing artificial neural networks. They investigate biases of NLP models with respect to hierarchical structure versus linear order by performing simulations involving simple recurrent networks (SRNs), gated recurrent units (GRUs), and long short-term memory units (LSTM). A crucial difference between the algorithms employed by this study to model language acquisition and the approach by the \textsc{modoma} relates to the levels of understanding an information-processing device defined by the seminal work of \citet[pp.~24-27]{Marr:1982} classifying algorithms at three levels: (1) computational theory, which is concerned with the goal, the logic, and the appropriateness of the procedure, (2) representation and algorithm, and (3) hardware implementation. In this respect, it could be argued that an artificial neural network provides an implementation at the level of hardware implementation while the \textsc{modoma} models language acquisition at the algorithmic level, which pertains to the representations of the input and output and the algorithm executing the transformation of input to output.

\section{Methods}
\label{methods}

The primary research question addressed in these experiments is: If we can replicate for \textsc{delilah}’s output the well-known finding that grammatical categories can be distinguished based on the linguistic contexts of words \citep[cf.][]{Schutze:1995}, can this finding be leveraged to let the \textsc{modoma} acquire grammatical categories and represent these categories by a rule-based grammar employing feature-value pairs? This question is explored through the following subquestions:

\begin{enumerate}[(1)]
\item \label{Q1} Considering the different types of clustering analyses, which clustering approach should be selected for this acquisition procedure?
\item \label{Q2} What number of groups should be configured as a parameter setting to acquire grammatical categories corresponding to more usual grammar descriptions, if applicable?
\item \label{Q3} Is it possible to use a data-mining analysis to acquire an abstract/discrete grammar?
\item \label{Q4} To what extent do the found groupings correspond to those suggested by conventional grammar models?
\item \label{Q5} Can the results obtained from the training experiment be replicated in the context of the test data?
\end{enumerate}

Taking into account previous studies performed with the \textsc{modoma} laboratory environment that indicated that only after being confronted with at least 10,000 exemplars, patterns were sufficiently revealed to the daughter to acquire discrete grammatical categories \citep[cf.][]{Shakouri:2025}, for this experiment only training and test sets containing 10,000 sentences generated by \textsc{delilah} were used. To carry out this experiment, the \textsc{modoma} employed R Statistics \citep{RCoreTeam:2020} to implement these analyses.\footnote{Amongst others, the R libraries cluster \citep{Maechler:2019}, plyr \citep{Wickham:2011,Wickham:2020a}, javaGD \citep{Urbanek:2012}, rJava \citep{Urbanek:2009,Urbanek:2017}, stat \citep{RCoreTeam:2020}, and tidyr (\citeauthor{Wickham:2016} \citeyear{Wickham:2016}, pp.~147--170; \citeauthor{Wickham:2020b} \citeyear{Wickham:2020b}) have been used for processing the data.} For this experiment a hierarchical agglomerative (bottom-up) clustering algorithm has been applied using the complete-link method, that is, data in a group are classified by taking into account the similarity with respect to the two most dissimilar data points. This technique has been selected based on experimentation by applying different algorithms on the training set with 10,000 utterances as the goal is to select the simplest algorithm that allows for the detection of syntactic categories. For example, for purposes of comparison experiments have been executed using divisive (top-down) clustering on the training data generated by \textsc{delilah} as well. As experiments using divisive clustering, which requires an additional processing step and may encounter issues with clusters assembling a few data points, revealed no significant differences compared to agglomerative clustering, a hierarchical bottom-up approach has been selected for the current experiment. Crucially, the use of clustering entails that the \textsc{modoma} implements an unsupervised algorithm, that is, no labels have been provided during the training phase by a linguist or another computer program such as the mother program or a tagger to indicate which classification training data have received. Thus, the groupings are only based on the characteristics of the dataset. This diverges from learning algorithms based on classification, for example \citet{Daelemans:2005} provide an example of an approach to learning based on classification rather than clustering.

The input to the clustering algorithm consists of n-grams with concordances representing keywords in context. Two \textsc{modoma} parameters specify the number of context items to be included respectively before and after the target word for further analysis. Currently, by default these parameters have been set to 2 context words before and after the key word, but depending on the experiment other settings could be used. A fragment of this type of list containing 57,733 exemplars is provided in Table \ref{tab:kwic} below. The translations and grammatical information regarding the target and context words have been provided solely for the convenience of the readers of this article, but these data have not been used as input to the \textsc{modoma} as this algorithm implements an unsupervised model of language acquisition.

The frequency of each combination of a target word and a context word is calculated considering its position relative to the target (e.g., as the second word before or directly preceding the target). The results for all target words are then combined into a single table, which lists the frequency of each target word and context word in a specific position. Words that are not found combined with a particular target word (for a position) but are attested for other target words, will still be listed for all key words in the table with a frequency of 0 for the unattested combinations. Using a \textsc{modoma} parameter, it is possible to specify that only data that occur in the dataset with a frequency above a threshold, should be included for further analyses. This table serves as input to the hierarchical agglomerative clustering algorithm and the clustering analysis performed is similar to the approach by \citet{Baayen:2006}. It is used to calculate a correlation matrix containing Spearman correlations. The result is squared to consider only the relative distances between correlations while not taking into account the differences between positive and negative correlations. Finally, the resulting correlation matrix is used to create a distance matrix indicating the relative distances between the different key words based on the context words. The outcome is used as input to the algorithm, which produces a hierarchical agglomerative clustering that can be visualized using a dendrogram (see for example Figure \ref{fig:dendr_training} below and Online Appendix 1 provided in the ancillary files). Based on manually inspecting the results of these analyses concerning the training data, it is possible to specify a number of discrete groups such that they combine the most similar words into clusters. Then, each of these groups will be considered as corresponding to a grammatical category and represented by a feature-value pair encoded by graph structures in the daughter language model's grammar. In particular, each item in the \textsc{modoma} daughter grammar with a phonological form corresponding to an item listed in one of these groups will be specified for the feature-value pair representing the corresponding grammatical category. Since these properties encode the combinatorial possibilities of lexical items, the result of acquisition has implications for unification (or merge) procedures involved in generating and parsing novel utterances.

Finally, to validate the parameter settings a second experiment is executed with respect to the test data containing 10,000 exemplars of the target language, which were independently generated by the mother language model. For this purpose, the same parameter settings are applied, which have been determined based on the simulation with the training data. To assess whether the experiment with the test data produces outcomes similar to those from the training data, we quantitatively evaluate the correspondence between the categories acquired by the daughter language model in both sessions. To this end, we apply Fisher's exact test to assess whether the association between the training and test clusters is significant. To analyze the results in more detail, post-hoc tests are performed using Fisher's exact test with Bonferroni correction to identify which categories contribute most to the association.

\begin{center}
\begin{table}
\hspace*{-1cm}													
\begin{tabularx}{\textwidth}{p{0.05\textwidth}p{0.19\textwidth}p{0.19\textwidth}p{0.19\textwidth}p{0.19\textwidth}p{0.19\textwidth}}
\cline{1-6}
\textbf{\#} & \centering\textbf{Context item 1} & \centering\textbf{Context item 2} & \centering\textbf{Target item} & \centering\textbf{Context item 3} & \centering\arraybackslash\textbf{Context item 4} \tabularnewline \cline{1-6}
1. & & & \centering\textit{werkt} & \centering\textit{niet} & \centering\arraybackslash\textit{elke} \\
& & & \centering{work:\textsc{pres:3:sg}} & \centering{not} & \centering\arraybackslash{every:\textsc{masc/fem}} \\ \cline{1-6}
2. & & \centering\textit{werkt} & \centering\textit{niet} & \centering\textit{elke} & \centering\arraybackslash\textit{fiets} \\
& & \centering{work:\textsc{pres:3:sg}} & \centering{not} & \centering{every:\textsc{masc/fem}} & \centering\arraybackslash{bike} \\ \cline{1-6}
3. & \centering\textit{werkt} & \centering\textit{niet} & \centering\textit{elke} & \centering\textit{fiets} & \\
& \centering{work:\textsc{pres:3:sg}} & \centering{not} & \centering{every:\textsc{masc/fem}} & \centering{bike} & \\ \cline{1-6}
4. & \centering\textit{niet} & \centering\textit{elke} & \centering\textit{fiets} & & \\
& \centering{not} & \centering{every:\textsc{masc/fem}} & \centering{bike} & & \\ \cline{1-6}
5. & & & \centering\textit{opnieuw} & \centering\textit{dient} & \centering\arraybackslash\textit{begeleiding} \\
& & & \centering{again} & \centering{serve:\textsc{pres:3:sg}} & \centering\arraybackslash{guidance} \\ \cline{1-6}
6. & & \centering\textit{opnieuw} & \centering\textit{dient} & \centering\textit{begeleiding} & \centering\arraybackslash\textit{te} \\
& & \centering{again} & \centering{serve:\textsc{pres:3:sg}} & \centering{guidance} & \centering\arraybackslash{to}\\ \cline{1-6}
7. & \centering\textit{opnieuw} & \centering\textit{dient} & \centering\textit{begeleiding} & \centering\textit{te} & \centering\arraybackslash\textit{zeggen} \\
& \centering{again} & \centering{serve:\textsc{pres:3:sg}} & \centering{guidance} & \centering{to}& \centering\arraybackslash{say:\textsc{inf}} \\ \cline{1-6}
8. & \centering\textit{dient} & \centering\textit{begeleiding} & \centering\textit{te} & \centering\textit{zeggen} & \centering\arraybackslash\textit{of} \\
& \centering{serve:\textsc{pres:3:sg}} & \centering{guidance} & \centering{to} & \centering{say:\textsc{inf}} & \centering\arraybackslash{whether} \\ \cline{1-6}
9. & \centering\textit{begeleiding} & \centering\textit{te} & \centering\textit{zeggen} & \centering\textit{of} & \centering\arraybackslash\textit{mooien} \\
& \centering{guidance} & \centering{to} & \centering{say:\textsc{inf}} & \centering{whether} & \centering\arraybackslash{pretty:\textsc{n:pl}} \\ \cline{1-6}
10. & \centering\textit{te} & \centering\textit{zeggen} & \centering\textit{of} & \centering\textit{mooien} & \centering\arraybackslash\textit{ineens} \\
& \centering{to} & \centering{say:\textsc{inf}} & \centering{whether} & \centering{pretty:\textsc{n:pl}} & \centering\arraybackslash{suddenly} \\ \cline{1-6}
11. & \centering\textit{zeggen} & \centering\textit{of} & \centering\textit{mooien} & \centering\textit{ineens} & \centering\arraybackslash\textit{zongen} \\
& \centering{say:\textsc{inf}} & \centering{whether} & \centering{pretty:\textsc{n:pl}} & \centering{suddenly} & \centering\arraybackslash{sing:\textsc{3:pl}} \\ \cline{1-6}
12. & \centering\textit{of} & \centering\textit{mooien} & \centering\textit{ineens} & \centering\textit{zongen} & \\
& \centering{whether} & \centering{pretty:\textsc{n:pl}} & \centering{suddenly} & \centering{sing:\textsc{3:pl}} & \\ \cline{1-6}
13. & \centering\textit{mooien} & \centering\textit{ineens} & \centering\textit{zongen} & & \\
& \centering{pretty:\textsc{n:pl}} & \centering{suddenly} & \centering{sing:\textsc{3:pl}} & & \\ \cline{1-6}
 \multicolumn{6}{c}{$\cdots$} \\ \cline{1-6}
57,731. & & \centering\textit{gaan} & \centering\textit{zijn} & \centering\textit{BMW’s} & \centering\arraybackslash\textit{lopen} \\
& & \centering{go:\textsc{pres:3:pl}} & \centering{his} & \centering{BMW:\textsc{pl}} & \centering\arraybackslash{walk:\textsc{inf}} \\ \cline{1-6}
57,732. & \centering\textit{gaan} & \centering\textit{zijn} & \centering\textit{BMW’s} & \centering\textit{lopen} & \\
& \centering{go:\textsc{pres:3:pl}} & \centering{his} & \centering{BMW:\textsc{pl}} & \centering{walk:\textsc{inf}} & \\ \cline{1-6}
57,733. & \centering\textit{zijn} & \centering\textit{BMW’s} & \centering\textit{lopen} & & \\
& \centering{his} & \centering{BMW:\textsc{pl}} & \centering{walk:\textsc{inf}} & & \\ \cline{1-6}
\end{tabularx}
\captionof{table}{Illustration of key words in context for use by further statistical analyses}
\label{tab:kwic}
\end{table}
\end{center}

\begin{landscape}
\begin{figure}[p]
\centering
\includegraphics[width=1.45\textwidth]{clustering_dendrogram_training.pdf}
\captionof{figure}{Structure of clustering based on 10,000 \textsc{delilah} exemplars for the data from session 1557861465468. For a full-size dendrogram presenting all data in a format that allows for detailed inspection of individual items, consult Online Appendix 1 provided in the ancillary files.}
\label{fig:dendr_training}
\end{figure}
\end{landscape}

\section{Results}
\label{results}

An analysis of the initial training dataset of 10,000 sentences generated by \textsc{delilah} revealed that the resulting clusters often aligned with grammatical categories commonly described in traditional grammars. The acquisition procedure was performed using the following parameter settings: a single run, exposure to a minimum of 10,000 utterances, a context window of two words before and two words after the target word, detection of 14 target clusters, and a minimum frequency threshold of 10 occurrences. The structure of the dendrogram resulting from the acquisition of grammatical categories based on the initial 10,000 sentences is presented in Figure \ref{fig:dendr_training} above. Online Appendix 1, available in the ancillary files, presents a full-size version of this extensive dendrogram with the data in a format that allows for easier inspection. To uniquely record and identify all data generated by the \textsc{modoma} system, each session is assigned a unique identification number based on the number of milliseconds elapsed since January 1, 1970. This dendrogram is a visualization of the numerical data the algorithm employs to acquire grammatical categories. It visualizes the respective (sub)groupings based on the contextual similarities between words and can be used to enhance the understanding of the acquired knowledge in relation to the dataset. Crucially, both the numerical data and the dendrogram represent an intermediate analysis of all data and the relative distances between words. Hence, while this analysis provides valuable insights, it does not lead to the identification of a discrete number of subgroups containing the most similar words that could correspond to a discrete number of categories. These categories are determined by an additional algorithm provided by R, which takes as its input the calculated numerical distances between all words.

Several observations can be made based on the result of this analysis. It lies beyond the scope of this article to provide an exhaustive listing. Nonetheless, the main clusters and several notable subgroupings will be discussed. Interestingly, this dendrogram shows that the third main left split groups many words that modify a noun, which would largely be classified as adjectives in traditional grammar. For the most part, the second main group from the left consists of words that modify a verb, namely adverbs. In the middle of the dendrogram, words that specify nouns, are grouped: Especially numerals are presented here as well as determiners such as articles, demonstrative pronouns, and indefinite pronouns. Within this cluster, there is also a subgroup of personal pronouns consisting of \textit{ik} (‘I’), \textit{jij} (‘you’, \textsc{inform}) and \textit{u} (‘you’, \textsc{formal}). Similarly, complementizers such as \textit{dat} (‘that’), \textit{terwijl} (‘while’) and \textit{nadat} (‘after’) are clustered and a subgroup that mainly consists of auxiliaries such as \textit{had} (‘had’, \textsc{sg}) and \textit{waren} (‘were’), can be discerned. Notably, prepositions such as \textit{met} (‘with’) and \textit{in} (‘in’) are grouped as well. Nonetheless, some words clustered with these prepositions would be classified differently according to traditional grammar (e.g., \textit{vergeten}, ‘forgotten’ and \textit{zeurt}, ‘nags’). Similarly, verbs such as \textit{huilen} (‘cry’) and \textit{bidden} (‘pray’) are grouped together. However, this verb cluster also includes words that are usually considered to belong to a different category (e.g., the adjectives \textit{actieve}, ‘active’ and \textit{onhandige}, ‘clumsy’). In the dendrogram in Figure \ref{fig:dendr_training}, perfect participles such as \textit{beroofd} (‘robbed’) and \textit{gemaakt} (‘made’) are clustered together. Finally, there is a cluster comprising various parts of speech according to more conventional grammar descriptions including nouns, verbs, participles, and complementizers. Interestingly, several subgroups in this cluster correspond to grammatical categories. For example, one subgroup contains prepositions such as \textit{af} (‘off’, ‘out’), \textit{over} (‘about’), \textit{voorbij} (‘past’), \textit{bij} (‘at’, ‘near’), and \textit{uit} (‘from’) while another subgroup consists of verbs, such as \textit{denken} (‘think’) and \textit{slapen} (‘sleep’). Another example is that \textit{daar} (‘there’) and \textit{hier} (‘here’) are clustered together. This suggests that the groupings indentified by the algorithm are non-trivial when compared to more conventional grammar models. 

\section{Analyses}
\label{analyses}

To be used to specify discrete categories, these hierarchical clusters should be divided into a number of groups. For the purposes of this experiment, 14 groups have been requested. The resulting clusters are listed in Table \ref{tab:groupings_training} below. For the convenience of non-Dutch speaking readers, some additional information on the meaning and grammatical properties of these lexical items is provided in Table \ref{tab:groupings_training} although these metadata have not been inputted into the unsupervised procedures by the language acquisition algorithm of the \textsc{modoma}. The contents of many clusters often appear to correspond closely with groupings made by more canonical accounts of grammar. For instance, cluster 13 only lists perfect participles. Interestingly, group 2 primarily consists of functional categories such as complementizers, prepositions and auxiliaries. Similarly, cluster 5 contains mainly words specifying nouns such as determiners (e.g., articles, demonstrative and indefinite pronouns, numerals, and personal pronouns). Groups 3 and 4 mostly list nouns while cluster 11 comprises nouns as well that are all but one diminutives. For the most part, clusters 1 and 12 consist of verbs and group 10 assembles some verbs and words related to \textit{vraag} (‘question’), that is, words taking a complement clause. Additionally, group 8 corresponds to sentential adverbs while groups 9 and 14 predominantly consist of adjectives. Conversely, cluster 6 alligns less with a category employed by traditional grammar descriptions as it contains for instance words that are usually classified as adjectives, adverbs, verbs, auxiliaries, and complementizers. Finally, group 7 coincides with diverse grammatical categories as well, but especially perfect participles and prepositions are represented.

Interestingly, the aim of the \textsc{modoma} is not to replicate the groupings established by conventional grammar accounts but to construct an abstracting model of the mother language. The goal of the daughter agent is to acquire at least a weakly equivalent grammar with respect to the mother language model. A weakly equivalent grammar considers the same set of sentences as grammatical as the mother agent. Weak equivalence is often contrasted with strong equivalence: A strongly equivalent grammar does not only allow for the same set of grammatical sentences but also assigns the same parses to them \citep[cf.][]{Chomsky:1963}. Thus, the weakly equivalent grammars are a subset of the strongly equivalent grammars and the acquisition of a strongly equivalent grammar would similarly result from successful acquisition procedures. In the case of the \textsc{modoma}, this would imply that after acquiring the target language the daughter language model can not only generate the same set of utterances as the mother language model but also assign the same parses in terms of the constituent structure to these utterances (i.e., it has acquired the same grammar except for label names).

Nonetheless, the correspondence with other descriptions of language proposed by linguists indicates that the daughter agent has acquired non-trivial knowledge of the target language. In this respect, it is noteworthy that the grammar \textsc{delilah}, that is, the mother language model in the \textsc{modoma} system, which produces the samples of language acquisition is based on, has been constructed by implementing an existing grammatical model and well-established analyses of Dutch. Therefore, while the acquired grammar for the present experiment does not result in a strongly equivalent model, this resemblance indicates the acquisition of a grammar that often assigns a similar parse with respect to the mother grammar. The acquisition of grammatical categories resembling those proposed by linguists can be seen as evidence of the soundness of the acquired grammar in terms of the equivalence between the mother and daughter language models. In this case the results of acquisition align with models derived from (often) centuries of linguistic research. Conversely, if different or diverging categories were acquired compared to those proposed by linguists, this would not imply that acquisition has led to a grammar that does not account for the target data. After all, the daughter is not requested to produce a strongly equivalent grammar and it is possible that a weakly equivalent model that employs completely different categories, could predict the same set of sentences. If so, this alternative analysis could provide insights for the field of linguistics by proposing new models. Nonetheless, such a result would need to be further explained.

\begin{center}
\begin{xltabular}{\textwidth}{ l l l X}
\hline \# & FEATURE & VALUE & Words in the cluster \\ \hline
1. & \textsc{a} & a & \textit{actieve} (‘active’), \textit{bidden} (‘pray’), \textit{blijken} (‘turn out’), \textit{concurreren} (‘compete’), \textit{gaan} (‘go’), \textit{groeien} (‘grow’), \textit{heb} (‘have’, \textsc{1:sg}), \textit{huilen} (‘cry’), \textit{hun} (‘their’), \textit{komen} (‘come’), \textit{lachen} (‘laugh’), \textit{onhandige} (‘clumsy’), \textit{overlijden} (‘decease’), \textit{praten} (‘talk’), \textit{'s} (‘of the’, \textsc{archaic det})\footnote{This archaic determiner was used by \textsc{delilah} as part of the collocation \textit{'s morgens}, ‘in the morning’.}, \textit{sociale} (‘social’), \textit{veranderen} (‘change’), \textit{verschijnen} (‘appear’), \textit{verschillen} (‘differ’), \textit{zwemmen} (‘swim’)\\ \hline
2. & \textsc{a} & b & \textit{aan} (‘to’, \textsc{prep}), \textit{als} (‘if’), \textit{alsof} (‘as if’), \textit{dan} (‘then’, ‘than’), \textit{dat} (‘that’), \textit{door} (‘by’), \textit{er} (‘there’), \textit{had} (‘had’, \textsc{sg}), \textit{hadden} (‘had’, \textsc{pl}), \textit{heeft} (‘has’), \textit{hoe} (‘how’), \textit{in} (‘in’), \textit{is} (‘is’), \textit{met} (‘with’), \textit{na} (‘after’), \textit{naar} (‘to’, \textsc{prep}), \textit{nadat} (‘after’), \textit{niet} (‘not’), \textit{of} (‘or’, ‘whether’), \textit{op} (‘on’), \textit{tegen} (‘against’), \textit{ter} (‘at’), \textit{terwijl} (‘while’), \textit{tot} (‘until’), \textit{totdat} (‘until’), \textit{van} (‘of’), \textit{vergeten} (‘forgotten’), \textit{voor} (‘for’), \textit{waar} (‘where’), \textit{waarom} (‘why’), \textit{wanneer} (‘when’), \textit{waren} (‘were’), \textit{was} (‘was’), \textit{werd} (‘became’, \textsc{sg}), \textit{werden} (‘became’, \textsc{pl}), \textit{wordt} (‘becomes’), \textit{zeurt} (‘nags’), \textit{zoals} (‘as’), \textit{zodat} (‘so that’)\\ \hline
3. & \textsc{a} & c & \textit{aansporingen} (‘incitements’), \textit{eigen} (‘own’), \textit{mogelijkheidjes} (‘small possibilities’), \textit{opdrachtjes} (‘small assignments’), \textit{ruimte} (‘space’), \textit{verwachting} (‘expectation’), \textit{zilveren} (‘silver’, \textsc{adj})\\ \hline
4. & \textsc{a} & d & \textit{aansporinkjes} (‘little incitements’), \textit{besluiten} (‘decisions’), \textit{besluitje} (‘little decision’), \textit{eis} (‘demand’, \textsc{n}), \textit{feiten} (‘facts’), \textit{kleine} (‘little’), \textit{mededelingen} (‘announcements’), \textit{onderwijs} (‘education’), \textit{opdrachten} (‘assignments’), \textit{overtuiginkjes} (‘little convictions’), \textit{samenhangende} (‘coherent’), \textit{uitwerking} (‘effect’), \textit{verwachtinkje} (‘small expectation’), \textit{voorsteltjes} (‘small proposals’), \textit{welzijn} (‘well being’, \textsc{n}) \\ \hline
5. & \textsc{a} & e & \textit{acht} (‘eight’), \textit{alle} (‘all’), \textit{blijkt} (‘turns out’), \textit{de} (‘the’, \textsc{masc/fem/pl}), \textit{deze} (‘this’, \textsc{masc/fem}), \textit{die} (‘that’, \textsc{masc/fem}), \textit{dit} (‘this’, \textsc{neut}), \textit{drie} (‘three’), \textit{een} (‘a’, ‘one’), \textit{elk} (‘every’, \textsc{neut}), \textit{elke} (‘every’, \textsc{masc/fem}), \textit{enkele} (‘a few’), \textit{geen} (‘no’), \textit{het} (‘the’, \textsc{neut:sg}), \textit{ieder} (‘every’, \textsc{neut}), \textit{iedere} (‘every’, \textsc{masc/fem}), \textit{ik} (‘I’), \textit{je} (‘you’, \textsc{inform}), \textit{jij} (‘you’, \textsc{inform}), \textit{massa’s} (‘masses’), \textit{menig} (‘many’, \textsc{sg}), \textit{menige} (‘many’, \textsc{pl}), \textit{negenennegentig} (‘ninetynine’), \textit{sommige} (‘some’, \textsc{pl}), \textit{te} (‘at’, ‘to’, \textsc{comp}), \textit{twaalf} (‘twelve’), \textit{twee} (‘two’), \textit{u} (‘you’, \textsc{formal}), \textit{veel} (‘a lot of’), \textit{vier} (‘four’), \textit{vijf} (‘five’), \textit{vijftig} (‘fifty’), \textit{weinig} (‘a few’), \textit{werkt} (‘works’)\\ \hline
6. & \textsc{a} & f & \textit{adequate} (‘adequate’), \textit{altijd} (‘always’), \textit{beseft} (‘realizes’), \textit{binnenlands} (‘domestic’), \textit{bleek} (‘seemed’), \textit{concurrentie} (‘competition’), \textit{daar} (‘there’), \textit{fossiele} (‘fossil’), \textit{gewenst} (‘desired’, \textsc{part}), \textit{hangen} (‘hang’), \textit{hebben} (‘have’), \textit{hebt} (‘have’, \textsc{2:sg}), \textit{hier} (‘here’), \textit{lacht} (‘laughs’), \textit{meer} (‘more’), \textit{meteen} (‘at once’), \textit{nooit} (‘never’), \textit{omdat} (‘because’), \textit{praat} (‘talks’), \textit{voordat} (‘before’), \textit{worden} (‘become’), \textit{zelfde} (‘same’), \textit{zijn} (‘are’, ‘be’, ‘his’)\\ \hline
7. & \textsc{a} & g & \textit{af} (‘off’, ‘out’), \textit{beschikking} (‘order’), \textit{bij} (‘at’, ‘near’), \textit{gebleken} (‘seemed’, \textsc{part}), \textit{gekomen} (‘come’, \textsc{part}), \textit{gekund} (‘able’, \textsc{past part} of \textit{kunnen}, ‘can’), \textit{gemoeten} (‘had to’, \textsc{part}), \textit{gevolgd} (‘followed’, \textsc{part}), \textit{gezeten} (‘sat’, \textsc{part}), \textit{gezien} (‘seen’), \textit{grove} (‘gross’), \textit{morgens} (‘of the morgen’, \textsc{archaic gen}), \textit{ontkend} (‘denied’, \textsc{part}), \textit{over} (‘about’), \textit{proberen} (‘try’), \textit{rond} (‘round’), \textit{terug} (‘back’), \textit{toe} (‘to’, ‘at’), \textit{toen} (‘then’, ‘when’), \textit{uit} (‘from’), \textit{voorbij} (‘past’), \textit{voort} (‘forth’), \textit{wereldwijde} (‘global’)\\ \hline
8. & \textsc{a} & h & \textit{al} (‘already’), \textit{bijna} (‘almost’), \textit{daarna} (‘thereafter’), \textit{daarom} (‘therefore’), \textit{echter} (‘however’), \textit{eerst} (‘at first’), \textit{eigenlijk} (‘actually’), \textit{even} (‘for a bit’), \textit{gaarne} (‘with pleasure’), \textit{gedeeltelijk} (‘partly’), \textit{ginds} (‘over there’), \textit{inderdaad} (‘indeed’), \textit{ineens} (‘suddenly’), \textit{misschien} (‘maybe’), \textit{momenteel} (‘at the moment’), \textit{natuurlijk} (‘of course’), \textit{nog} (‘yet’), \textit{nu} (‘now’), \textit{ook} (‘also’), \textit{opnieuw} (‘again’), \textit{soms} (‘sometimes’), \textit{steeds} (‘still’), \textit{tegelijkertijd} (‘at the same time’), \textit{thans} (‘at present’), \textit{toch} (‘however’), \textit{verder} (‘further’), \textit{voorheen} (‘previously’), \textit{waarschijnlijk} (‘probably’), \textit{weer} (‘again’), \textit{wel} (‘well’), \textit{zo} (‘so’, ‘immediately’)\\ \hline
9. & \textsc{a} & i & \textit{arrogante} (‘arrogant’), \textit{behandeling} (‘treatment’), \textit{beschikbare} (‘available’), \textit{binnenlandse} (‘domestic’), \textit{blije} (‘happy’), \textit{boze} (‘angry’), \textit{bruto} (‘gross’), \textit{creatieve} (‘creative’), \textit{culturele} (‘cultural’), \textit{demografische} (‘demographic’), \textit{directe} (‘direct’), \textit{dramatische} (‘dramatic’), \textit{ethische} (‘ethical’), \textit{eventuele} (‘possible’), \textit{financiele} (‘financial’), \textit{functionele} (‘functional’), \textit{fundamentele} (‘fundamental’), \textit{gedachtetjes} (‘little thoughts’), \textit{gehele} (‘complete’), \textit{gewone} (‘normal’), \textit{goede} (‘good’), \textit{haar} (‘her’), \textit{hele} (‘whole’), \textit{hobbelige} (‘bumpy’), \textit{invoering} (‘introduction’), \textit{leerplichtige} (‘in compulsary education’), \textit{medische} (‘medical’), \textit{minderjarige} (‘underage’), \textit{museale} (‘museological’), \textit{nieuwe} (‘new’), \textit{officiele} (‘official’), \textit{onschatbare} (‘invaluable’), \textit{onschuldige} (‘innocent’), \textit{onwerkelijke} (‘surreal’), \textit{Oost-Europese} (‘Eastern European’), \textit{positieve} (‘positive’), \textit{prachtige} (‘beautiful’), \textit{psychische} (‘psychic’), \textit{ruimtelijke} (‘spatial’), \textit{schaarse} (‘scarce’), \textit{sociaal-economische} (‘socio-economic’), \textit{speciale} (‘special’), \textit{spectaculaire} (‘spectacular’), \textit{stapsgewijze} (‘gradual’), \textit{strafbare} (‘punishable’), \textit{vaste} (‘fixed’), \textit{verre} (‘remote’), \textit{volledige} (‘complete’), \textit{voortdurende} (‘ongoing’), \textit{voortvarende} (‘vigorous’), \textit{waardevolle} (‘valuable’), \textit{zichtbare} (‘visible’)\\ \hline
10. & \textsc{a} & j & \textit{begrijpen} (‘understand’), \textit{horen} (‘hear’), \textit{lezende} (‘reading’), \textit{merken} (‘notice’), \textit{vraag} (‘question’), \textit{vraagje} (‘small question’), \textit{vraagjes} (‘small questions’), \textit{vragen} (‘questions’, \textsc{n}, ‘question’, \textsc{v}), \textit{weten} (‘know’), \textit{zien} (‘see’)\\ \hline
11. & \textsc{a} & k & \textit{behandelingen} (‘treatments’), \textit{behandelingetjes} (‘small treatments’), \textit{beloftetjes} (‘little promises’), \textit{beschrijvinkjes} (‘small descriptions’), \textit{besluitjes} (‘small decisions’), \textit{bestralinkjes} (‘small irradiations’), \textit{eisjes} (‘small demands’), \textit{impulsjes} (‘small impulses’), \textit{kansjes} (‘small opportunities’), \textit{operaties} (‘operations’, ‘surgeries’), \textit{stimulansjes} (‘small incentives’), \textit{vermogentjes} (‘small abilities’, ‘small assets’), \textit{verwachtinkjes} (‘small expectations’)\\ \hline
12. & \textsc{a} & l & \textit{behoren} (‘belong’), \textit{beseffen} (‘realize’), \textit{blijkende} (‘turning out’), \textit{denken} (‘think’), \textit{dienen} (‘serve’), \textit{durven} (‘dare’), \textit{gillen} (‘scream’), \textit{kleven} (‘stick’), \textit{lijken} (‘seem’), \textit{om} (‘to’), \textit{ontkennen} (‘deny’), \textit{slapen} (‘sleep’), \textit{snijden} (‘cut’), \textit{stijgen} (‘rise’), \textit{tevens} (‘also’), \textit{wensen} (‘wish’), \textit{wensende} (‘wishing’), \textit{werken} (‘work’), \textit{winkelen} (‘shop’), \textit{zitten} (‘sit’), \textit{zittende} (‘sitting’)\\ \hline
13. & \textsc{a} & m & \textit{beroofd} (‘robbed’, \textsc{part}), \textit{betreurd} (‘regretted’, \textsc{part}), \textit{geaccepteerd} (‘accepted’, \textsc{part}), \textit{geboden} (‘offered’, \textsc{part}), \textit{gebouwd} (‘built’, \textsc{part}), \textit{gehoord} (‘heard’, \textsc{part}), \textit{geleerd} (‘learned’, \textsc{part}), \textit{gemaakt} (‘made’, \textsc{part}), \textit{gemerkt} (‘noticed’, \textsc{part}), \textit{gesproken} (‘spoken’, \textsc{part}), \textit{gevonden} (‘found’, \textsc{part}), \textit{gevraagd} (‘asked’, \textsc{part}), \textit{gezegd} (‘said’, \textsc{part}), \textit{onderstreept} (‘underlines’, ‘underlined’, \textsc{part}), \textit{vereisende} (‘demanding’), \textit{voorkomen} (‘prevent’, ‘prevented’, \textsc{part}), \textit{zeggen} (‘say’)\\ \hline
14. & \textsc{a} & n & \textit{internationale} (‘international’), \textit{jonge} (‘young’), \textit{meervoudige} (‘multiple’), \textit{Nederlandse} (‘Dutch’), \textit{omvangrijke} (‘extensive’), \textit{raad} (‘advice’), \textit{trage} (‘slow’)\\ \hline
\end{xltabular}
\captionof{table}{Groupings resulting from the cluster analysis of the data from session 1557861465468}
\label{tab:groupings_training}
\end{center}

Each grouping in Table \ref{tab:groupings_training} is assigned a feature-value pair representing a grammatical category. These feature-value pairs are subsequently used to specify all items in the grammar of the daughter language model that have a phonological representation corresponding to the words listed in this table for each category. Crucially, all results of cluster analysis are represented using the same feature, in this case \textsc{a}. However, depending on the cluster, the items are assigned different values, ranging from \textit{a} to \textit{n}. The feature (i.e., \textsc{a}) can be interpreted as corresponding to a major linguistic category such as part of speech while the values (i.e., \textit{a} to \textit{n}) represent specific parts of speech (e.g., noun, adjective, adverb, or verb, as per conventional grammatical descriptions) that have been acquired. For instance, adverbs will be subcategorized as $\langle\mbox{\textsc{a}:h}\rangle$ whereas most adjectives will be classified as either $\langle\mbox{\textsc{a}:i}\rangle$ or $\langle\mbox{\textsc{a}:n}\rangle$ as both clusters 9 and 14 mainly list adjectives.

From a more formal perspective, the feature can be regarded as a dimension with the specific classification representing a value within that dimension: Each requested language acquisition procedure introduces a major classification of linguistic knowledge such as part of speech or grammatical number into the grammar employed by the daughter language model. Conversely, the specific results of language acquisistion indicate which grammatical knowledge has been acquired with respect to this major category, for example, noun or second person. Then, each requested acquisition simulation either introduces a new type of linguistic knowledge to the grammar such as functional vs. content categories for a grammar that is already specifying part of speech and number or adds additional information with respect to previously acquired grammatical classifications. For instance, a new learning procedure could also model the acquisition of grammatical knowledge related to part of speech: In this case, the acquired categories should also be classified as values within this major classification, namely feature \textsc{a}. Conversely, if subsequent acquisition procedures are designed to acquire knowledge of another type of grammatical classification such as the distinction between function and content words, a different feature (e.g., \textsc{b}) should be used to specify the results in the daughter grammar. Then, the values for this feature --- such as \textit{a} and \textit{b} representing function word vs. content word --- correspond to the specific categories that could be acquired. 

Crucially, in the \textsc{modoma} daughter language model formalism, a graph structure encoding a grammatical item can only be indicated for a single value for each feature or major linguistic category. Conversely, it can simultaneously be specified for multiple grammatical features. For instance, the lemma for \textit{onderwijs} (‘education’) can be specified as $\langle\mbox{\textsc{a}:d}\rangle$ (corresponding to $\langle\mbox{\textsc{partofspeech}:noun}\rangle$) and $\langle\mbox{\textsc{b}:b}\rangle$ (e.g., indicating $\langle\mbox{\textsc{functionorcontent}:content}\rangle$) simultaneously. To the contrary, it is impossible to classify this lemma as both $\langle\mbox{\textsc{a}:a}\rangle$, which corresponds to $\langle\mbox{\textsc{partofspeech}:verb}\rangle$ and $\langle\mbox{\textsc{a}:d}\rangle$, which is associated with the grammatical category indication $\langle\mbox{\textsc{partofspeech}:verb}\rangle$. Thus, this formalism enables the specification of a language model in which grammatical items are explicitly defined in terms of the acquired features while preventing contradictory classifications. This example illustrates the added value of using feature-value pairs rather than single labels to indicate acquired grammatical categories.

Moreover, by encoding the results of acquisition by feature-value pairs in the grammar, these properties can be used as restrictions on the combinatory properties of words and constructions during parsing and generating new utterances. Thus, the daughter agent has acquired categories it can use to generate grammatical utterances (according to the current grammar) and determine whether utterances generated by the mother language model are grammatical. In accordance with the interaction model, that is, acquisition is based on an ongoing interaction between the mother and daughter agents rather than a previously assembled corpus, the \textsc{modoma} system is designed in such a way that as soon as new grammatical knowledge has been acquired, it is employed to take part in the exchanging of sentences with the mother. It is a noteworthy characteristic of the \textsc{modoma} that unlike many existing language acquisition algorithms knowledge is used before the grammar has been fully acquired. As a result, the utterances produced by the daughter language model based on newly acquired grammatical items are instantly input to feedback by the mother language model, if requested by the users of the system. Thus, an assessment of the quality of the newly acquired grammatical knowledge can be carried out by the daughter language model.

\section{Evaluating the Suggested Default Settings on Another Dataset}
\label{eval_set}

A second experiment was conducted using a test set of 10,000 sentences. The sentences were independently generated during a separate session by \textsc{delilah} employing the same settings as those determined by the training data. Thus, the proposed parameter settings were validated on an independent dataset. The structure of the results of hierarchical agglomerative cluster analysis on this test set containing 10,000 \textsc{delilah} utterances are presented by Figure \ref{fig:dendr_test} below. A full-size version of these results, which facilitates a more detailed inspection of individual data items and their relationships, is provided in Online Appendix 2 in the ancillary files. We will highlight some major and noteworthy groupings identified in this analysis. However, this account is not exhaustive. In this dendrogram, the first major split groups sentential adverbs to the left. This clustering of a specific subclass of adverbs is a non-trivial result. It is followed in the graph to the right by a group of words that take a complement clause. Most of them are infinitive or third person plural verbs such as \textit{horen}, ‘hear’, and \textit{zeggen}, ‘say’, except for the present participle \textit{lezende} (‘reading’) and two nouns related to \textit{vraag} (‘question’, \textit{vraagje}, ‘small question’). There are two subgroups that assemble nouns. Interestingly, the plural word forms connected to \textit{vraag}, ‘question’, that is, \textit{vragen}, a nominal meaning ‘questions’, which ambiguously also includes the verbal form \textit{vragen}, ‘question’, and \textit{vraagjes}, ‘small questions’, are grouped with the nouns here.

Moreover, there is a major split that primarily groups determiners, pronouns, and numerals. The first subcluster within this split also contains two verbs: \textit{blijkt}, (‘turns out’) and \textit{werkt}, (‘works’). It is insightful that a subgroup assembles the personal pronouns \textit{je} (‘you’ \textsc{clit.inform}), \textit{jij} (‘you’, \textsc{inform}), and \textit{u} (‘you’, \textsc{formal}). Similarly, this cluster includes numerals and determiners such as \textit{acht} (‘eight’), \textit{twee} (‘two’), and \textit{sommige} (‘some’, \textsc{pl}) grouped together. Interestingly, the dendrogram in Figure \ref{fig:dendr_test} reveals that \textit{massa’s} (‘masses’) is grouped with the numerals and this correctly detects that the generator of \textsc{delilah} uses this word similarly to numerals. Thus, analysis by the \textsc{modoma} can increase knowledge of the results of processing by \textsc{delilah}. Within this cluster, numerals and determiners that are not numerals such as \textit{die} (‘that’, \textsc{masc/fem}), \textit{deze} (‘this’, \textsc{masc/fem}), \textit{de} (‘the’, \textsc{masc/fem/pl}), \textit{het} (‘the’, \textsc{neut.sg}), and \textit{geen} (‘no’), are placed together as well. Another insightful observation is that \textit{elk} (‘every’, \textsc{neut}) and \textit{ieder} (‘every’, \textsc{neut}) are grouped together in a subcluster as well as \textit{elke} (‘every’, \textsc{masc/fem}) and \textit{iedere} (‘every’, \textsc{masc/fem}).

 In the dendrogram in Figure \ref{fig:dendr_test}, there is a cluster that assembles functional categories. Moreover, the first subcluster in this group of functional categories includes only subjunctive conjunctions (e.g., \textit{totdat}, ‘until’, \textit{wanneer}, ‘when’, and \textit{waarom}, ‘why’) while the next subgroup primarily lists prepositions such as \textit{aan} (‘to’), \textit{in} (‘in’), and \textit{met} (‘with’). Thus, these groupings provide valuable insights and are indicative of grammatical structures at multiple levels. A noteworthy major split consists of nominals, mainly adjectives such as \textit{internationale} (‘international’), \textit{vervelende} (‘annoying’), and \textit{warme} (‘warm’) along with some nouns (e.g., \textit{vrede}, ‘peace’, and \textit{ethiek}, ‘ethics’) while another major cluster largely contains verbs. There are several subgroups within this major verb cluster: One subgroup consists especially of third person verbs such as \textit{groeit} (‘grows’) and \textit{functioneert} (‘functions’). Two exceptions are the adjectives \textit{kortdurende} (‘short-term’) and \textit{sociaal-economische} (‘socio-economic’). Moreover, a subgroup within the verb cluster assembles in particular various verbal forms such as third and plural person auxiliaries, for instance, \textit{wordt}, ‘becomes’, and \textit{worden}, ‘become’, plural forms (e.g., \textit{huilden}, ‘cried’, \textsc{pl}), and past perfect participles (e.g., \textit{gehouden}, ‘held’, and \textit{gekomen}, ‘come’). Notably, the infinitive complementizer \textit{te} (‘to’) is also grouped with the verbal forms. Interestingly, within this group there are subclusters of auxiliaries: One such subgroup contains \textit{worden}, ‘become’, \textit{hebben}, ‘have’, and \textit{zijn}, ‘are’, ‘be’, which is synonymous with the word for ‘his’, while another cluster includes \textit{heeft}, ‘has’, and \textit{waren}, ‘were’. Within the latter, \textit{is}, ‘is’, and \textit{was}, ‘was’, are grouped together with a small distance to \textit{waren}, ‘were’ in the dendrogram.

Similarly to the results of the hierarchical agglomerative clustering analysis of the training set, the test set containing 10,000 sentences also reveals a major cluster that corresponds less clearly to grammatical categories traditionally used in grammar descriptions. Nonetheless, several subgroups within this cluster primarily contain words that align with categories described in conventional grammar such as the groupings of \textit{daar} (‘there’) and \textit{hier} (‘here’), the prepositions \textit{over} (‘about’) and \textit{uit} (‘from’), and the possessives \textit{mijn} (‘my’) and \textit{uw} (‘your’, \textsc{formal}). Another interesting observation is that there is a main split that largely contains infinite verb forms in an ordered manner. For instance, a subgroup within this cluster specifically includes perfect past participles (e.g., \textit{gedaan}, ‘done’, and \textit{geboden}, ‘offered’). Other clusters in this major split predominantly group verbal forms as well such as \textit{zittende} (‘sitting’), \textit{dienen} (‘serve’), and \textit{zitten} (‘sit’) while the following main split consists of third person plural or infinitive verbs including \textit{beseffen} (‘realize’), \textit{hangen} (‘hang’), and \textit{willen} (‘want’). Finally, a major split should be noted, which again encompasses nominals: These are mostly adjectives such as \textit{zware} (‘heavy’), \textit{onhandige} (‘clumsy’), and \textit{algemene} (‘general’) as well as some nouns (e.g., \textit{aanbod}, ‘offer’, \textit{verwachtinkje}, ‘small expectation’, and \textit{koffie}, ‘coffee’).

\begin{landscape}
\begin{figure}[p]
\centering
\includegraphics[width=1.45\textwidth]{clustering_dendrogram_test.pdf}
\captionof{figure}{Structure of clustering based on 10,000 \textsc{delilah} exemplars for the data from session 1601581107338. For a full-size dendrogram presenting all data in a format that allows for detailed inspection of individual items, consult Online Appendix 2 provided in the ancillary files.}
\label{fig:dendr_test}
\end{figure}
\end{landscape}

Taking the results of the intermediate analysis based on hierarchical agglomerative clustering analysis, 14 grammatical categories have been requested employing the parameter settings determined by the first session. Table \ref{tab:groupings_test} presents the resulting categories. The translations and grammatical details with respect to these words have only been provided to facilitate the interpretation of the results but were not available to the unsupervised language processing procedures by the \textsc{modoma}. Interestingly, the second category in Table \ref{tab:groupings_test} contains functional categories, which consist of prepositions (e.g., \textit{aan}, ‘to’, and \textit{voor}, ‘for’), subjunctive complementizers (e.g., \textit{dat}, ‘that’, and \textit{omdat}, ‘because’), and the auxiliary \textit{hadden} (‘had’, \textsc{pl}). Similarly, with the exception of the verb \textit{werkt} (‘works’) category 6 corresponds to determiners. Categories 3, 7 and 9 primarily include nominals (i.e., adjectives as well as nouns) while categories 4, 5 and 11 only list nouns. Group 14 consists exclusively of verbs and cluster 13 contains mostly verbs. Additionally, category 8 includes mainly past participles along with some prepositions. Category 12 lists words that take a sentential complement, namely verbs and words connected to \textit{vraag} (‘question’). Finally, category 10 contains sentential adverbs. Conversely, the first group does not seem to correspond clearly to a category described by conventional grammars as it contains for instance adverbs such as \textit{daar} (‘there’) and \textit{dan} (‘then’, also meaning ‘than’), adjectives (e.g., \textit{binnenlands}, ‘domestic’, and \textit{sociaal-economische}, ‘socio-economic’), various verbal forms (e.g., \textit{functioneert}, ‘functions’, and \textit{zijn}, ‘are’, ‘be’, which is synonymous with the word for ‘his’), possessives such as \textit{mijn} (‘my’), and prepositions (e.g., \textit{over}, ‘about’, and \textit{uit}, ‘from’).

\begin{center}
\begin{xltabular}{\textwidth}{ l l l X}
\hline \# & FEATURE & VALUE & Words in the cluster \\ \hline
1. & \textsc{a} & a & \textit{ben} (‘am’), \textit{binnenlands} (‘domestic’), \textit{daar} (‘there’), \textit{dan} (‘then’, ‘than’), \textit{directe} (‘direct’), \textit{er} (‘there’), \textit{ernstige} (‘serious’), \textit{functioneert} (‘functions’), \textit{gaan} (‘go’), \textit{gaat} (‘goes’), \textit{gehouden} (‘held’, \textsc{part}), \textit{gekomen} (‘come’, \textsc{part}), \textit{getoond} (‘shown’), \textit{groeit} (‘grows’), \textit{had} (‘had’, \textsc{sg}), \textit{hangt} (‘hangs’), \textit{hebben} (‘have’), \textit{hebt} (‘have’, \textsc{2:sg}), \textit{heeft} (‘has’), \textit{hier} (‘here’), \textit{huilden} (‘cried’, \textsc{pl}), \textit{hun} (‘their’), \textit{is} (‘is’), \textit{komt} (‘comes’), \textit{kortdurende} (‘short-term’), \textit{kwaadaardige} (‘evil’), \textit{meer} (‘more’), \textit{mijn} (‘my’), \textit{morgens} (‘of the morgen’, \textsc{archaic gen}), \textit{niet} (‘not’), \textit{over} (‘about’), \textit{'s} (‘of the’, \textsc{archaic det}), \textit{sociaal-economische} (‘socio-economic’), \textit{speelt} (‘plays’), \textit{te} (‘at’, ‘to’, \textsc{comp}), \textit{ter} (‘at’), \textit{terug} (‘back’), \textit{uit} (‘from’), \textit{uw} (‘your’, \textsc{formal}), \textit{verandert} (‘changes’), \textit{vereisende} (‘demanding’), \textit{verschijnt} (‘appears’), \textit{waren} (‘were’), \textit{was} (‘was’), \textit{wensende} (‘wishing’), \textit{werd} (‘became’, \textsc{sg}), \textit{werden} (‘became’, \textsc{pl}), \textit{worden} (‘become’), \textit{wordt} (‘becomes’), \textit{zijn} (‘are’, ‘be’, ‘his’)\\ \hline
2. & \textsc{a} & b & \textit{aan} (‘to’, \textsc{prep}), \textit{als} (‘if’), \textit{alsof} (‘as if’), \textit{bij} (‘at’, ‘near’), \textit{dat} (‘that’), \textit{door} (‘by’), \textit{hadden} (‘had’, \textsc{pl}), \textit{hoe} (‘how’), \textit{in} (‘in’), \textit{met} (‘with’), \textit{naar} (‘to’, \textsc{prep}), \textit{of} (‘or’, ‘whether’), \textit{omdat} (‘because’), \textit{op} (‘on’), \textit{tegen} (‘against’), \textit{terwijl} (‘while’), \textit{tot} (‘until’), \textit{totdat} (‘until’), \textit{van} (‘of’), \textit{vergeten} (‘forgotten’), \textit{voor} (‘for’), \textit{voordat} (‘before’), \textit{waar} (‘where’), \textit{waarom} (‘why’), \textit{wanneer} (‘when’), \textit{zoals} (‘as’), \textit{zodat} (‘so that’)\\ \hline
3. & \textsc{a} & c & \textit{aanbod} (‘offer’), \textit{actieve} (‘active’), \textit{dramatische} (‘dramatic’), \textit{gezamenlijke} (‘common’), \textit{langdurige} (‘long-lasting’), \textit{luie} (‘lazy’), \textit{medische} (‘medical’), \textit{museale} (‘museological’), \textit{nationale} (‘national’), \textit{omvangrijke} (‘extensive’), \textit{onderwijs} (‘education’), \textit{schone} (‘clean’), \textit{verwachtinkje} (‘small expectation’), \textit{zichtbare} (‘visible’), \textit{zware} (‘heavy’)\\ \hline
4. & \textsc{a} & d & \textit{aansporing} (‘incitement’), \textit{beloftetjes} (‘little promises’), \textit{besluitjes} (‘small decisions’), \textit{eisjes} (‘small demands’), \textit{initiatiefjes} (‘little initiatives’), \textit{kans} (‘opportunity’), \textit{kansen} (‘opportunities’), \textit{mededelingen} (‘announcements’), \textit{mededelingetjes} (‘little announcements’), \textit{mogelijkheden} (‘possibilities’), \textit{mogelijkheidjes} (‘small possibilities’), \textit{opdracht} (‘assignment’), \textit{opdrachten} (‘assignments’), \textit{verwachtinkjes} (‘small expectations’), \textit{voorstellen} (‘proposals’), \textit{voorsteltjes} (‘small proposals’) \\ \hline
5. & \textsc{a} & e & \textit{aansporingen} (‘incitements’), \textit{feitjes} (‘small facts’), \textit{kansjes} (‘small opportunities’), \textit{vraagjes} (‘small questions’), \textit{vragen} (‘questions’, \textsc{n}, ‘question’, \textsc{v}) \\ \hline
6. & \textsc{a} & f & \textit{acht} (‘eight’), \textit{alle} (‘all’), \textit{blijkt} (‘turns out’), \textit{de} (‘the’, \textsc{masc/fem/pl}), \textit{deze} (‘this’, \textsc{masc/fem}), \textit{die} (‘that’, \textsc{masc/fem}), \textit{dit} (‘this’, \textsc{neut}), \textit{drie} (‘three’), \textit{een} (‘a’, ‘one’), \textit{elk} (‘every’, \textsc{neut}), \textit{elke} (‘every’, \textsc{masc/fem}), \textit{enkele} (‘a few’), \textit{geen} (‘no’), \textit{het} (‘the’, \textsc{neut:sg}), \textit{ieder} (‘every’, \textsc{neut}), \textit{iedere} (‘every’, \textsc{masc/fem}), \textit{ik} (‘I’), \textit{je} (‘you’, \textsc{inform}), \textit{jij} (‘you’, \textsc{inform}), \textit{massa’s} (‘masses’), \textit{menig} (‘many’, \textsc{sg}), \textit{menige} (‘many’, \textsc{pl}), \textit{negenennegentig} (‘ninetynine’), \textit{sommige} (‘some’, \textsc{pl}), \textit{twaalf} (‘twelve’), \textit{twee} (‘two’), \textit{u} (‘you’, \textsc{formal}), \textit{veel} (‘a lot of’), \textit{vier} (‘four’), \textit{vijf} (‘five’), \textit{vijftig} (‘fifty’), \textit{weinig} (‘a few’), \textit{werkt} (‘works’)\\ \hline
7. & \textsc{a} & g & \textit{adequate} (‘adequate’), \textit{algemene} (‘general’), \textit{arme} (‘poor’), \textit{belangrijkere} (‘more important’), \textit{dankbare} (‘grateful’), \textit{etnische} (‘ethnic’), \textit{goede} (‘good’), \textit{grijze} (‘grey’), \textit{haar} (‘her’), \textit{halve} (‘half’), \textit{initiatieven} (‘initiatives’), \textit{koffie} (‘coffee’), \textit{lange} (‘lang’), \textit{multi-etnische} (‘multi-ethnic’), \textit{nadrukkelijke} (‘explicit’), \textit{onhandige} (‘clumsy’), \textit{slaap} (‘sleep’), \textit{spelen} (‘play’), \textit{vitale} (‘vital’), \textit{WAO} (‘disability law’)\\ \hline
8. & \textsc{a} & h & \textit{af} (‘off’, ‘out’), \textit{begrepen} (‘understood’, \textsc{part}), \textit{betreurd} (‘regretted’, \textsc{part}), \textit{bevorderd} (‘promoted’, \textsc{part}), \textit{gebleken} (‘seemed’, \textsc{part}), \textit{geboden} (‘offered’, \textsc{part}), \textit{gebouwd} (‘built’, \textsc{part}), \textit{gedaan} (‘done’, \textsc{part}), \textit{gehoord} (‘heard’, \textsc{part}), \textit{gekund} (‘able’, \textsc{past part} of \textit{kunnen}, ‘can’), \textit{gemoeten} (‘had to’, \textsc{part}), \textit{gezegd} (‘said’, \textsc{part}), \textit{onderstreept} (‘underlines’, ‘underlined’, \textsc{part}), \textit{ontkend} (‘denied’, \textsc{part}), \textit{toe} (‘to’, ‘at’), \textit{verantwoord} (‘responsible’), \textit{voorkomen} (‘prevent’, ‘prevented’, \textsc{part}), \textit{voort} (‘forth’)\\ \hline
9. & \textsc{a} & i & \textit{afschuwelijke} (‘horrible’), \textit{agressieve} (‘aggressive’), \textit{arbeidsongeschikte} (‘disabled’), \textit{binnenlandse} (‘domestic’), \textit{boze} (‘angry’), \textit{duurzame} (‘durable’), \textit{ethiek} (‘ethics’), \textit{ethische} (‘ethical’), \textit{functionele} (‘functional’), \textit{gezonde} (‘healthy’), \textit{grove} (‘gross’), \textit{internationale} (‘international’), \textit{langzame} (‘slow’), \textit{last} (‘burden’), \textit{lekkere} (‘tasty’), \textit{minderjarige} (‘underage’), \textit{Nederlandse} (‘Dutch’), \textit{ongekende} (‘unprecedented’), \textit{onmisbare} (‘indispensible’), \textit{openbare} (‘public’), \textit{preventieve} (‘preventive’), \textit{rare} (‘strange’), \textit{relaxte} (‘relaxt’), \textit{rustige} (‘quiet’), \textit{second} (‘second’)\footnote{This is an English loanword, which was used by \textsc{delilah} as part of the collocation \textit{second opinion}.}, \textit{snelle} (‘quick’), \textit{stapsgewijze} (‘gradual’), \textit{uitwerking} (‘effect’), \textit{veilige} (‘safe’), \textit{verkeerde} (‘wrong’), \textit{verre} (‘remote’), \textit{vervelende} (‘annoying’), \textit{vrede} (‘peace’), \textit{warme} (‘warm’), \textit{zelfde} (‘same’), \textit{zieke} (‘ill’)\\ \hline
10. & \textsc{a} & j & \textit{al} (‘already’), \textit{altijd} (‘always’), \textit{bijna} (‘almost’), \textit{binnenkort} (‘soon’), \textit{daarna} (‘thereafter’), \textit{daarom} (‘therefore’), \textit{echter} (‘however’), \textit{eerst} (‘at first’), \textit{eigenlijk} (‘actually’), \textit{even} (‘for a bit’), \textit{gaarne} (‘with pleasure’), \textit{gedeeltelijk} (‘partly’), \textit{ginds} (‘over there’), \textit{inderdaad} (‘indeed’), \textit{ineens} (‘suddenly’), \textit{meteen} (‘at once’), \textit{misschien} (‘maybe’), \textit{momenteel} (‘at the moment’), \textit{natuurlijk} (‘of course’), \textit{nog} (‘yet’), \textit{nooit} (‘never’), \textit{nu} (‘now’), \textit{ook} (‘also’), \textit{opnieuw} (‘again’), \textit{soms} (‘sometimes’), \textit{tegelijkertijd} (‘at the same time’), \textit{tevens} (‘also’), \textit{thans} (‘at present’), \textit{toch} (‘however’), \textit{toen} (‘then’, ‘when’), \textit{verder} (‘further’), \textit{vooral} (‘especially’), \textit{voorheen} (‘previously’), \textit{waarschijnlijk} (‘probably’), \textit{weer} (‘again’), \textit{wel} (‘well’), \textit{zo} (‘so’, ‘immediately’)\\ \hline
11. & \textsc{a} & k & \textit{amputatietjes} (‘small amputations’), \textit{behandelingen} (‘treatments’), \textit{beschrijvingen} (‘descriptions’), \textit{bestralingen} (‘irradiations’), \textit{bestralinkjes} (‘small irradiations’), \textit{operaties} (‘operations’, ‘surgeries’), \textit{operatietjes} (‘small operations’, ‘small surgeries’)\\ \hline
12. & \textsc{a} & l & \textit{begrijpen} (‘understand’), \textit{horen} (‘hear’), \textit{lezen} (‘read’), \textit{lezende} (‘reading’), \textit{merken} (‘notice’), \textit{vraag} (‘question’), \textit{vraagje} (‘small question’), \textit{weten} (‘know’), \textit{zeggen} (‘say’), \textit{zien} (‘see’)\\ \hline
13. & \textsc{a} & m & \textit{behoren} (‘belong’), \textit{beseft} (‘realizes’), \textit{blijkende} (‘turning out’), \textit{dienen} (‘serve’), \textit{durven} (‘dare’), \textit{gewenst} (‘desired’, \textsc{part}), \textit{om} (‘to’), \textit{proberen} (‘try’), \textit{verlangen} (‘desire’), \textit{vermogens} (‘abilities’, ‘assets’), \textit{vermogentjes} (‘small abilities’, ‘small assets’), \textit{zitten} (‘sit’), \textit{zittende} (‘sitting’)\\ \hline
14. & \textsc{a} & n & \textit{beseffen} (‘realize’), \textit{bidden} (‘pray’), \textit{blijken} (‘turn out’), \textit{concurreren} (‘compete’), \textit{denken} (‘think’), \textit{drentelen} (‘saunter’), \textit{gebeuren} (‘happen’), \textit{groeien} (‘grow’), \textit{hangen} (‘hang’), \textit{huilen} (‘cry’), \textit{kleven} (‘stick’), \textit{komen} (‘come’), \textit{liggen} (‘lie’), \textit{lijken} (‘seem’), \textit{ontkennen} (‘deny’), \textit{praten} (‘talk’), \textit{raden} (‘guess’), \textit{slapen} (‘sleep’), \textit{snijden} (‘cut’), \textit{staan} (‘stand’), \textit{stijgen} (‘rise’), \textit{veranderen} (‘change’), \textit{verbeteren} (‘improve’), \textit{wensen} (‘wish’), \textit{werken} (‘work’), \textit{willen} (‘want’), \textit{winkelen} (‘shop’), \textit{zwemmen} (‘swim’)\\ \hline
\end{xltabular}
\captionof{table}{Groupings resulting from the cluster analysis of the data from session 1601581107338}
\label{tab:groupings_test}
\end{center}

Many categories in Table \ref{tab:groupings_test} appear to correspond to the results of the previous experiment, which involved performing the same analysis with the same parameter settings on 10,000 training sentences generated by \textsc{delilah} as well. Not only are clusters corresponding to part of speech categories typically used in conventional grammar models such as noun, verb, and sentential adverb identified, but other grammatical categories are also distinguished for both datasets. For instance, the second group identified in the previous experiment corresponds to functional categories such as prepositions, complementizers, and a form of the auxiliary \textit{hebben} (‘have’). Similarly, for both the training and test datasets categories containing verbs and the word \textit{vraag} (‘question’), which take a sentential complement, as well as groups seemingly not corresponding to a category employed by more traditional grammars are discovered.

To quantitatively assess the correspondence between the categories resulting from the training and test sessions, we have created the crosstabulation presented in Table \ref{tab:crosstabs_training_test}, which displays the number of overlapping lexical items for each possible combination of training and test categories. Subsequently, we have used Fisher's exact test with 500,000 Monte Carlo simulations to determine whether the association between the training and test clusters is significant at $p < 0.05$. As this analysis resulted in p = $2 \times 10^{-6}$, the null hypothesis of no association between the training and test categories is rejected. To further assess which training and test categories contribute most to this association, pairwise Fisher's exact tests with Bonferroni-correction were carried out using R and the RVAideMemoire package \citep{Herve:2022} for each two-by-two subtable that is entailed by Table \ref{tab:crosstabs_training_test}, see the study by \citet{MacDonald:2000} for a similar approach employing pairwise chi-square tests. These post-hoc tests indicated that in particular the subtables containing acquired training and test categories with overlapping lexical items contributed to this significant result. For example, Fisher's exact tests with Bonferroni-correction performed on the two-by-two clusters for training categories $\langle\mbox{A:b}\rangle$ and $\langle\mbox{A:h}\rangle$ and test categories $\langle\mbox{A:b}\rangle$ and $\langle\mbox{A:j}\rangle$ as well as on training categories $\langle\mbox{A:e}\rangle$ and $\langle\mbox{A:j}\rangle$ and test categories $\langle\mbox{A:f}\rangle$ and $\langle\mbox{A:l}\rangle$ resulted in significant p-values of respectively p = $1.328 \times 10^{-11}$ and p = $8.667 \times 10^{-5}$. This analysis is consistent with the observation that categories acquired during both experiments appear to correspond to each other.

\begin{table}
\centering
\adjustbox{valign=c, width=\textwidth}{ 				
\begin{tabular} {l c c c c c c c c c c c c c c }
\hline & \multicolumn{14}{c}{\bf{Training}} \\ \cline{2-15}
\bf{Test} & A:a & A:b & A:c & A:d & A:e & A:f & A:g & A:h & A:i & A:j & A:k & A:l & A:m & A:n \\ \hline
A:a & 3 & 12 & 0 & 0 & 1 & 8 & 5 & 0 & 2 & 0 & 0 & 1 & 1 & 0 \\
A:b & 0 & 23 & 0 & 0 & 0 & 2 & 1 & 0 & 0 & 0 & 0 & 0 & 0 & 0 \\
A:c & 1 & 0 & 0 & 2 & 0 & 0 & 0 & 0 & 4 & 0 & 0 & 0 & 0 & 1 \\
A:d & 0 & 0 & 1 & 3 & 0 & 0 & 0 & 0 & 0 & 0 & 4 & 0 & 0 & 0 \\
A:e & 0 & 0 & 1 & 0 & 0 & 0 & 0 & 0 & 0 & 2 & 1 & 0 & 0 & 0 \\
A:f & 0 & 0 & 0 & 0 & 33 & 0 & 0 & 0 & 0 & 0 & 0 & 0 & 0 & 0 \\
A:g & 1 & 0 & 0 & 0 & 0 & 1 & 0 & 0 & 2 & 0 & 0 & 0 & 0 & 0 \\
A:h & 0 & 0 & 0 & 0 & 0 & 0 & 7 & 0 & 0 & 0 & 0 & 0 & 7 & 0 \\
A:i & 0 & 0 & 0 & 1 & 0 & 1 & 1 & 0 & 7 & 0 & 0 & 0 & 0 & 2 \\
A:j & 0 & 0 & 0 & 0 & 0 & 3 & 1 & 30 & 0 & 0 & 0 & 1 & 0 & 0 \\
A:k & 0 & 0 & 0 & 0 & 0 & 0 & 0 & 0 & 0 & 0 & 3 & 0 & 0 & 0 \\
A:l & 0 & 0 & 0 & 0 & 0 & 0 & 0 & 0 & 0 & 8 & 0 & 0 & 1 & 0 \\
A:m & 0 & 0 & 0 & 0 & 0 & 2 & 1 & 0 & 0 & 0 & 1 & 7 & 0 & 0 \\
A:n & 9 & 0 & 0 & 0 & 0 & 1 & 0 & 0 & 0 & 0 & 0 & 11 & 0 & 0 \\ \hline 
\end{tabular}
}
\captionof{table}{Crosstabulation of the overlap between the acquired categories during the training and test experiments}
\label{tab:crosstabs_training_test}
\end{table}

\section{Conclusions and Suggestions for Future Research}
\label{conclusions}
This article presents experiments conducted to evaluate the \textsc{modoma} system. The \textsc{modoma} serves as a multi-agent computational laboratory environment for language acquisition simulations enabling controlled experimentation and analysis. It employs two language models: a mother agent and a daughter agent. Both language models utilize comprehensive representations of linguistic knowledge for example including grammatical categories as well as the phonological and semantic forms of lexical items. The adult agent is based on \textsc{Delilah}. This is a language model, which has been independently developed by \citet{Cremers:1995} and provides a generator and parser executing a grammar model of Dutch \citep[cf.][for a detailed discussion of this language model]{Cremers:2014}. Conversely, the daughter agent is a novel language model, which has been specifically developed for the \textsc{modoma}. It employs a language acquisition device to acquire knowledge of the target language. The result is used to define a grammar model of the linguistic knowledge of the target language, which is explicitly encoded through graph structures. These graphs employ feature-value pairs to represent the acquired linguistic categories. Moreover, the daughter language model includes a generator and a parser, which can execute her currently acquired grammar. Thus, language acquisition results in a productive language model. Crucially, as the \textsc{modoma} provides a laboratory environment for language acquisition experiments, all factors involved in language acquisition such as the adult agent, the language acquiring agent, and the exchange of utterances are integral components in a single system. Accordingly, all aspects of the system can be configured by the user through parameter settings while all results are logged and can be retrieved. This design opens up new possibilities for further research into language acquisition. Building on the work by \citet{Shakouri:2025}, the experiments presented in this article serve to both evaluate and illustrate the potential of this design.

In comparison with large language models, this line of research provides an interesting additional perspective of language modelling. A major advantage of a computational language acquisition laboratory along the lines of the \textsc{modoma} is that it offers inherently transparent artificial intelligence agents: The explicit representations of grammar entail that all knowledge employed by the system to process language can be consulted. Moreover, as all components involved in language acquisition including the adult and child agents are part of the simulation, the system implements a fully accountable model of first language acquisition. This creates the possibility of conducting experiments that would be impossible to perform using human subjects such as comparing diverse experimental conditions. Given the growing use of language models, studying the extent to which machine-generated language resembles human language patterns is becoming increasingly significant. The \textsc{modoma} explores this issue, which is exemplified by this study. Finally, a notable design aspect is that the daughter agent only employs unsupervised techniques to construct a model of the target language. This resembles human children acquiring their mother language, who do not receive direct information on the labels employed by the adults. Thus, this system expands the perspective on modelling language acquisition.

This article discusses the results of two experiments aimed at the acquisition of grammatical categories for instance consisting of nouns, verbs, or determiners employing cluster analysis. The first experiment was conducted to determine the parameter settings while the second was used to validate these settings. The specification of discrete graph-based representations through unsupervised methods presents significant computational challenges. Therefore, a hybrid approach was employed in both experiments involving statistical as well as rule-based techniques. The statistical technique selected for this analysis was hierarchical agglomerative clustering, which was applied to a dataset consisting of 10,000 sentences generated by mother language model, see Figure \ref{fig:dendr_training} and Online Appendix 1 in the ancillary files for the intermediate results of the first experiment. This analysis resulted in a continuous representation of the grammatical relationships between the words. However, the daughter language model is designed to explicitly represent discrete grammatical categories. Therefore, based on the results of the training experiment, 14 categories were defined grouping the most similar items. Using these selected parameter settings, the first experiment demonstrated the feasibility of modelling the acquisition of discrete grammatical categories. These categories were represented using feature-value pairs specifying graph structures and the result was used to update the grammar model employed by the daughter language model. Therefore, this analysis enabled the system to acquire an abstract discrete grammar model of the target language. In particular, similarly to how grammatical categories are employed by more conventional grammar models, these updated structures indicate restrictions on the grammaticality of combinations of linguistic structures. Thus, these categories can be used productively by a language model employing a parser and/or a generator to generate grammatical structures and evaluate the grammaticality of input utterances depending on the currently acquired grammar. Non-trivially, the resulting categories largely correspond to distinctions made by more traditional grammatical models constructed by linguists, as shown in Table \ref{tab:groupings_training}.

Using the default settings from the initial experiment, a new experiment was conducted to analyze 10,000 sentence utterances generated by the mother language model. Hence, hierarchical agglomerative clustering should yield 14 discrete categories for this second experiment as well. The intermediate results of this second simulation are displayed in Figure \ref{fig:dendr_test} and Online Appendix 2 in the ancillary files while the resulting 14 categories are presented in Table \ref{tab:groupings_test}. Similarly to the first experiment, these categories often correspond to those used in conventional grammar models. This further suggests that the methods and parameter settings used to identify the patterns and acquire the categories are non-trivial. In both experiments, the \textsc{modoma} successfully specifies a discrete language model that describes abstractions similar to those found in other accounts of language. Concomitantly, statistical analyses of the overlap in lexical items between the training and test categories (see Table \ref{tab:crosstabs_training_test}) revealed a significant association between the categories acquired as a result of both experiments at $p < 0.05$. Post-hoc testing further indicated that this association is primarily driven by the training and test categories with matching lexical items. Accordingly, these analyses substantiate the observation that the acquired categories identified in both experiments are related. Thus, using the same parameter settings, the results from the training experiment are successfully replicated in the context of the test data.

Summarizing, by taking a statistical analysis as an intermediate step the acquisition procedures resulted in an abstracting rule-based grammar that encodes grammatical knowledge by leveraging discrete categories represented through feature-value pairs. Moreover, as soon as this abstracting grammatical knowledge has been acquired, it is applied by the daughter language model in two ways:
\begin{enumerate}[(1)]
\item The newly acquired knowledge is recorded in the grammar by specifying the lemmas with phonological forms identified by the learning procedure as corresponding to an acquired category. These lemmas are subcategorized for these feature-value pairs, which impose constraints on the combinatory properties of the lexical entries during parsing and generation. Therefore, the results of acquisition determine the grammaticality of newly parsed and produced utterances. Crucially, this final step entails the acquisition of abstract rule-based grammatical knowledge, which enables the daughter language model to make grammaticality judgements.
\item The daughter language immediately utilizes the updated lemmas to exchange utterances with the mother agent. 
\end{enumerate}

Crucially, this study demonstrated that the \textsc{modoma} model of first language acquisition enables the acquisition of discrete grammatical categories and this knowledge is respresented by the daughter language model in such a way that it can be used productively to generate and parse novel utterances. A major contribution of this study is that it validates the concept of computational modeling in first language acquisition through the use of multi-agent systems: The experiments presented in this article strongly suggest the feasibility of advancing research into the computational modelling of language acquisition using multi-agent systems and offer a basis for multiple future research possibilities in this area. For example, the results of acquisition can be evaluated based on feedback, if requested. This suggests an interesting direction for further exploration. Moreover, previously acquired grammatical knowledge can be used as input to subsequent acquisition procedures. This learning strategy has been called internal annotation in the context of the \textsc{modoma}, which is a type of self-supervised learning \citep[e.g.,][]{Balestriero:2023, Gui:2024}. The \textsc{modoma} laboratory environment is an all-encompassing system, which includes all aspects of language acquisition from the generation of the samples of the target language by the mother and statistical analyses to adding newly acquired knowledge to the grammar and executing successive and diverse language acquisition strategies. Therefore, the use of internal annotation is an effective strategy for a \textsc{modoma}: In addition to employing newly acquired grammatical rules to generate and parse sentences, acquired grammatical knowledge can also be used to enable further acquisition procedures. The findings presented in this article establish a foundation for further exploration of internal annotation, with a particular focus on a multi-agent model of first language acquisition. These aspects will be the focus of upcoming studies.

\section*{Acknowledgements}
\label{acknowledgements}
The authors wish to thank Maarten Hijzelendoorn for his help and advice.
This publication is part of the \textsc{modoma} project, which was financed by the Dutch Research Council (NWO).

\bibliographystyle{apacite}
\bibliography{Shakouri_et_al_art_unsup_acq_dic_cat}

\providecommand{\noopsort}[1]{}
\begin{thebibliography}{}

\bibitem [\protect \citeauthoryear {%
Alishahi%
\ \BBA {} Chrupa\l{}a%
}{%
Alishahi%
\ \BBA {} Chrupa\l{}a%
}{%
{\protect \APACyear {2009}}%
}]{%
Alishahi:2009}
\APACinsertmetastar {%
Alishahi:2009}%
\begin{APACrefauthors}%
Alishahi, A.%
\BCBT {}\ \BBA {} Chrupa\l{}a, G.%
\end{APACrefauthors}%
\unskip\
\newblock
\APACrefYearMonthDay{2009}{}{}.
\newblock
{\BBOQ}\APACrefatitle {Lexical category acquisition as an incremental process}
  {Lexical category acquisition as an incremental process}.{\BBCQ}
\newblock
\BIn{} \APACrefbtitle {Proceedings of the {C}og{S}ci 2009 Workshop on {P}sycho
  {C}omputational {M}odels of {H}uman {L}anguage {A}cquisition.} {Proceedings
  of the {C}og{S}ci 2009 workshop on {P}sycho {C}omputational {M}odels of
  {H}uman {L}anguage {A}cquisition.}
\PrintBackRefs{\CurrentBib}

\bibitem [\protect \citeauthoryear {%
Alishahi%
\ \BBA {} Chrupa\l{}a%
}{%
Alishahi%
\ \BBA {} Chrupa\l{}a%
}{%
{\protect \APACyear {2012}}%
}]{%
Alishahi:2012}
\APACinsertmetastar {%
Alishahi:2012}%
\begin{APACrefauthors}%
Alishahi, A.%
\BCBT {}\ \BBA {} Chrupa\l{}a, G.%
\end{APACrefauthors}%
\unskip\
\newblock
\APACrefYearMonthDay{2012}{}{}.
\newblock
{\BBOQ}\APACrefatitle {Concurrent acquisition of word meaning and lexical
  categories} {Concurrent acquisition of word meaning and lexical
  categories}.{\BBCQ}
\newblock
\BIn{} \APACrefbtitle {Proceedings of the 2012 {J}oint {C}onference on
  {E}mpirical {M}ethods in {N}atural {L}anguage {P}rocessing and
  {C}omputational {N}atural {L}anguage {L}earning} {Proceedings of the 2012
  {J}oint {C}onference on {E}mpirical {M}ethods in {N}atural {L}anguage
  {P}rocessing and {C}omputational {N}atural {L}anguage {L}earning}\ (\BPGS\
  643--654).
\newblock
\APACaddressPublisher{}{Association for Computational Linguistics (ACL)}.
\newblock
\begin{APACrefURL} \url{https://aclanthology.org/D12-1059/} \end{APACrefURL}
\PrintBackRefs{\CurrentBib}

\bibitem [\protect \citeauthoryear {%
Alishahi%
\ \BBA {} Stevenson%
}{%
Alishahi%
\ \BBA {} Stevenson%
}{%
{\protect \APACyear {2008}}%
}]{%
Alishahi:2008}
\APACinsertmetastar {%
Alishahi:2008}%
\begin{APACrefauthors}%
Alishahi, A.%
\BCBT {}\ \BBA {} Stevenson, S.%
\end{APACrefauthors}%
\unskip\
\newblock
\APACrefYearMonthDay{2008}{}{}.
\newblock
{\BBOQ}\APACrefatitle {A computational model of early argument structure
  acquisition} {A computational model of early argument structure
  acquisition}.{\BBCQ}
\newblock
\APACjournalVolNumPages{Cognitive Science}{32}{5}{789--834}.
\newblock
\begin{APACrefDOI} \doi{https://doi.org/10.1080/03640210801929287}
  \end{APACrefDOI}
\PrintBackRefs{\CurrentBib}

\bibitem [\protect \citeauthoryear {%
Baayen%
, Feldman%
\BCBL {}\ \BBA {} Schreuder%
}{%
Baayen%
\ \protect \BOthers {.}}{%
{\protect \APACyear {2006}}%
}]{%
Baayen:2006}
\APACinsertmetastar {%
Baayen:2006}%
\begin{APACrefauthors}%
Baayen, R\BPBI H.%
, Feldman, L\BPBI B.%
\BCBL {}\ \BBA {} Schreuder, R.%
\end{APACrefauthors}%
\unskip\
\newblock
\APACrefYearMonthDay{2006}{}{}.
\newblock
{\BBOQ}\APACrefatitle {Morphological influences on the recognition of
  monosyllabic monomorphemic words} {Morphological influences on the
  recognition of monosyllabic monomorphemic words}.{\BBCQ}
\newblock
\APACjournalVolNumPages{Journal of Memory and Language}{55}{2}{290--313}.
\newblock
\begin{APACrefDOI} \doi{https://doi.org/10.1016/j.jml.2006.03.008}
  \end{APACrefDOI}
\PrintBackRefs{\CurrentBib}

\bibitem [\protect \citeauthoryear {%
Baldridge%
\ \BBA {} Kruijff%
}{%
Baldridge%
\ \BBA {} Kruijff%
}{%
{\protect \APACyear {2003}}%
}]{%
Baldridge:2003}
\APACinsertmetastar {%
Baldridge:2003}%
\begin{APACrefauthors}%
Baldridge, J.%
\BCBT {}\ \BBA {} Kruijff, G\BHBI J\BPBI M.%
\end{APACrefauthors}%
\unskip\
\newblock
\APACrefYearMonthDay{2003}{}{}.
\newblock
{\BBOQ}\APACrefatitle {Multi-modal combinatory categorial grammar} {Multi-modal
  combinatory categorial grammar}.{\BBCQ}
\newblock
\BIn{} A.~Copestake\ \BBA {} J.~Haji{\v{c}}\ (\BEDS), \APACrefbtitle {10th
  {C}onference of the {E}uropean {C}hapter of the {A}ssociation for
  {C}omputational {L}inguistics ({EACL})} {10th {C}onference of the {E}uropean
  {C}hapter of the {A}ssociation for {C}omputational {L}inguistics ({EACL})}\
  (\BPGS\ 211--218).
\newblock
\APACaddressPublisher{}{Association for Computational Linguistics (ACL)}.
\newblock
\begin{APACrefURL} \url{https://aclanthology.org/E03-1036} \end{APACrefURL}
\PrintBackRefs{\CurrentBib}

\bibitem [\protect \citeauthoryear {%
Balestriero%
\ \protect \BOthers {.}}{%
Balestriero%
\ \protect \BOthers {.}}{%
{\protect \APACyear {2023}}%
}]{%
Balestriero:2023}
\APACinsertmetastar {%
Balestriero:2023}%
\begin{APACrefauthors}%
Balestriero, R.%
, Ibrahim, M.%
, Sobal, V.%
, Morcos, A.%
, Shekhar, S.%
, Goldstein, T.%
\BDBL {}Goldblum, M.%
\end{APACrefauthors}%
\unskip\
\newblock
\APACrefYearMonthDay{2023}{}{}.
\newblock
{\BBOQ}\APACrefatitle {A cookbook of self-supervised learning} {A cookbook of
  self-supervised learning}.{\BBCQ}
\newblock
\APACjournalVolNumPages{Computing Research Repository
  (CoRR)}{{arXiv:cs.LG/2304.12210v2}}{}{}.
\newblock
\begin{APACrefDOI} \doi{https://doi.org/10.48550/arXiv.2304.12210}
  \end{APACrefDOI}
\PrintBackRefs{\CurrentBib}

\bibitem [\protect \citeauthoryear {%
Beekhuizen%
, Bod%
, Fazly%
, Stevenson%
\BCBL {}\ \BBA {} Verhagen%
}{%
Beekhuizen%
\ \protect \BOthers {.}}{%
{\protect \APACyear {2014}}%
}]{%
Beekhuizen:2014}
\APACinsertmetastar {%
Beekhuizen:2014}%
\begin{APACrefauthors}%
Beekhuizen, B.%
, Bod, R.%
, Fazly, A.%
, Stevenson, S.%
\BCBL {}\ \BBA {} Verhagen, A.%
\end{APACrefauthors}%
\unskip\
\newblock
\APACrefYearMonthDay{2014}{}{}.
\newblock
{\BBOQ}\APACrefatitle {A usage-based model of early grammatical development} {A
  usage-based model of early grammatical development}.{\BBCQ}
\newblock
\BIn{} V.~Demberg\ \BBA {} T.~O'Donell\ (\BEDS), \APACrefbtitle {Proceedings of
  the {F}ifth {W}orkshop on {C}ognitive {M}odeling and {C}omputational
  {L}inguistics {(CMCL)}} {Proceedings of the {F}ifth {W}orkshop on {C}ognitive
  {M}odeling and {C}omputational {L}inguistics {(CMCL)}}\ (\BPGS\ 46--54).
\newblock
\APACaddressPublisher{}{Association for Computational Linguistics (ACL)}.
\newblock
\begin{APACrefDOI} \doi{https://doi.org/10.3115/v1/W14-2006} \end{APACrefDOI}
\PrintBackRefs{\CurrentBib}

\bibitem [\protect \citeauthoryear {%
Bod%
, Scha%
\BCBL {}\ \BBA {} Sima’an%
}{%
Bod%
\ \protect \BOthers {.}}{%
{\protect \APACyear {2003}}%
}]{%
Bod:2003}
\APACinsertmetastar {%
Bod:2003}%
\begin{APACrefauthors}%
Bod, R.%
, Scha, R.%
\BCBL {}\ \BBA {} Sima’an, K.%
\end{APACrefauthors}%
\unskip\
\newblock
\APACrefYear{2003}.
\newblock
\APACrefbtitle {Data-oriented parsing} {Data-oriented parsing}.
\newblock
\APACaddressPublisher{Stanford, CA}{CSLI}.
\PrintBackRefs{\CurrentBib}

\bibitem [\protect \citeauthoryear {%
Brown%
\ \protect \BOthers {.}}{%
Brown%
\ \protect \BOthers {.}}{%
{\protect \APACyear {2020}}%
}]{%
Brown:2020}
\APACinsertmetastar {%
Brown:2020}%
\begin{APACrefauthors}%
Brown, T\BPBI B.%
, Mann, B.%
, Ryder, N.%
, Subbiah, M.%
, Kaplan, J.%
, Dhariwal, P.%
\BDBL {}Amodei, D.%
\end{APACrefauthors}%
\unskip\
\newblock
\APACrefYearMonthDay{2020}{}{}.
\newblock
{\BBOQ}\APACrefatitle {Language models are few-shot learners} {Language models
  are few-shot learners}.{\BBCQ}
\newblock
\BIn{} H.~Larochelle, M.~Ranzato, R.~Hadsell, M\BHBI F.~Balcan\BCBL {}\ \BBA {}
  H\BHBI T.~Lin\ (\BEDS), \APACrefbtitle {Advances in neural information
  processing systems 33 ({NeurIPS} 2020)} {Advances in neural information
  processing systems 33 ({NeurIPS} 2020)}\ (\BVOL~33, \BPGS\ 1877--1901).
\newblock
\APACaddressPublisher{}{Curran Associates}.
\newblock
\begin{APACrefURL}
  \url{https://papers.nips.cc/paper\_files/paper/2020/file/1457c0d6bfcb4967418bfb8ac142f64a-Paper.pdf}
  \end{APACrefURL}
\PrintBackRefs{\CurrentBib}

\bibitem [\protect \citeauthoryear {%
Chaabouni%
, Kharitonov%
, Bouchacourt%
, Dupoux%
\BCBL {}\ \BBA {} Baroni%
}{%
Chaabouni%
\ \protect \BOthers {.}}{%
{\protect \APACyear {2020}}%
}]{%
Chaabouni:2020}
\APACinsertmetastar {%
Chaabouni:2020}%
\begin{APACrefauthors}%
Chaabouni, R.%
, Kharitonov, E.%
, Bouchacourt, D.%
, Dupoux, E.%
\BCBL {}\ \BBA {} Baroni, M.%
\end{APACrefauthors}%
\unskip\
\newblock
\APACrefYearMonthDay{2020}{}{}.
\newblock
{\BBOQ}\APACrefatitle {Compositionality and generalization in emergent
  languages} {Compositionality and generalization in emergent
  languages}.{\BBCQ}
\newblock
\BIn{} D.~Jurafsky, J.~Chai, N.~Schluter\BCBL {}\ \BBA {} J.~Tetreault\
  (\BEDS), \APACrefbtitle {{P}roceedings of the 58th {A}nnual {M}eeting of the
  {A}ssociation for {C}omputational {L}inguistics} {{P}roceedings of the 58th
  {A}nnual {M}eeting of the {A}ssociation for {C}omputational {L}inguistics}\
  (\BPGS\ 4427--4442).
\newblock
\APACaddressPublisher{}{Association for Computational Linguistics (ACL)}.
\newblock
\begin{APACrefDOI} \doi{https://doi.org/10.18653/v1/2020.acl-main.407}
  \end{APACrefDOI}
\PrintBackRefs{\CurrentBib}

\bibitem [\protect \citeauthoryear {%
Chaabouni%
, Kharitonov%
, Dupoux%
\BCBL {}\ \BBA {} Baroni%
}{%
Chaabouni%
\ \protect \BOthers {.}}{%
{\protect \APACyear {2019}}%
}]{%
Chaabouni:2019}
\APACinsertmetastar {%
Chaabouni:2019}%
\begin{APACrefauthors}%
Chaabouni, R.%
, Kharitonov, E.%
, Dupoux, E.%
\BCBL {}\ \BBA {} Baroni, M.%
\end{APACrefauthors}%
\unskip\
\newblock
\APACrefYearMonthDay{2019}{}{}.
\newblock
{\BBOQ}\APACrefatitle {Anti-efficient encoding in emergent communication}
  {Anti-efficient encoding in emergent communication}.{\BBCQ}
\newblock
\BIn{} H\BPBI M.~Wallach, H.~Larochelle, A.~Beygelzimer, F.~d'Alch\'{e} Buc,
  E\BPBI B.~Fox\BCBL {}\ \BBA {} R.~Garnett\ (\BEDS), \APACrefbtitle
  {{A}dvances in {N}eural {I}nformation {P}rocessing {S}ystems 32 ({NeurIPS}
  2019)} {{A}dvances in {N}eural {I}nformation {P}rocessing {S}ystems 32
  ({NeurIPS} 2019)}\ (\BVOL~32, \BPGS\ 6261--6271).
\newblock
\APACaddressPublisher{}{Curran Associates}.
\newblock
\begin{APACrefURL}
  \url{https://proceedings.neurips.cc/paper\_files/paper/2019/file/31ca0ca71184bbdb3de7b20a51e88e90-Paper.pdf}
  \end{APACrefURL}
\PrintBackRefs{\CurrentBib}

\bibitem [\protect \citeauthoryear {%
Chaabouni%
, Kharitonov%
, Dupoux%
\BCBL {}\ \BBA {} Baroni%
}{%
Chaabouni%
\ \protect \BOthers {.}}{%
{\protect \APACyear {2021}}%
}]{%
Chaabouni:2021}
\APACinsertmetastar {%
Chaabouni:2021}%
\begin{APACrefauthors}%
Chaabouni, R.%
, Kharitonov, E.%
, Dupoux, E.%
\BCBL {}\ \BBA {} Baroni, M.%
\end{APACrefauthors}%
\unskip\
\newblock
\APACrefYearMonthDay{2021}{}{}.
\newblock
{\BBOQ}\APACrefatitle {Communicating artificial neural networks develop
  efficient color-naming systems} {Communicating artificial neural networks
  develop efficient color-naming systems}.{\BBCQ}
\newblock
\APACjournalVolNumPages{{P}roceedings of the {N}ational {A}cademy of {S}ciences
  ({PNAS})}{118}{12}{}.
\newblock
\begin{APACrefDOI} \doi{https://doi.org/10.1073/pnas.2016569118}
  \end{APACrefDOI}
\PrintBackRefs{\CurrentBib}

\bibitem [\protect \citeauthoryear {%
Chaabouni%
\ \protect \BOthers {.}}{%
Chaabouni%
\ \protect \BOthers {.}}{%
{\protect \APACyear {2022}}%
}]{%
Chaabouni:2022}
\APACinsertmetastar {%
Chaabouni:2022}%
\begin{APACrefauthors}%
Chaabouni, R.%
, Strub, F.%
, Altch\'{e}, F.%
, Tallec, C.%
, Trassov, E.%
, Davoodi, E.%
\BDBL {}Piot, B.%
\end{APACrefauthors}%
\unskip\
\newblock
\APACrefYearMonthDay{2022}{}{}.
\newblock
{\BBOQ}\APACrefatitle {Emergent communication at scale} {Emergent communication
  at scale}.{\BBCQ}
\newblock
\BIn{} \APACrefbtitle {{P}roceedings of the {T}enth {I}nternational
  {C}onference on {L}earning {R}epresentations ({ICLR} 2022).} {{P}roceedings
  of the {T}enth {I}nternational {C}onference on {L}earning {R}epresentations
  ({ICLR} 2022).}
\newblock
\begin{APACrefURL} \url{https://openreview.net/pdf?id=AUGBfDIV9rL}
  \end{APACrefURL}
\PrintBackRefs{\CurrentBib}

\bibitem [\protect \citeauthoryear {%
Chomsky%
}{%
Chomsky%
}{%
{\protect \APACyear {1963}}%
}]{%
Chomsky:1963}
\APACinsertmetastar {%
Chomsky:1963}%
\begin{APACrefauthors}%
Chomsky, N.%
\end{APACrefauthors}%
\unskip\
\newblock
\APACrefYearMonthDay{1963}{}{}.
\newblock
{\BBOQ}\APACrefatitle {Formal properties of grammar} {Formal properties of
  grammar}.{\BBCQ}
\newblock
\BIn{} R\BPBI D.~Luce, R\BPBI R.~Bush\BCBL {}\ \BBA {} E.~Galanter\ (\BEDS),
  \APACrefbtitle {Handbook of Mathematical Psychology} {Handbook of
  mathematical psychology}\ (\BVOL~II, \BPGS\ 323--418).
\newblock
\APACaddressPublisher{New York, NY}{Wiley}.
\PrintBackRefs{\CurrentBib}

\bibitem [\protect \citeauthoryear {%
Conner%
, Gertner%
, Fisher%
\BCBL {}\ \BBA {} Roth%
}{%
Conner%
\ \protect \BOthers {.}}{%
{\protect \APACyear {2008}}%
}]{%
Conner:2008}
\APACinsertmetastar {%
Conner:2008}%
\begin{APACrefauthors}%
Conner, M.%
, Gertner, Y.%
, Fisher, C.%
\BCBL {}\ \BBA {} Roth, D.%
\end{APACrefauthors}%
\unskip\
\newblock
\APACrefYearMonthDay{2008}{}{}.
\newblock
{\BBOQ}\APACrefatitle {Baby {SRL}: Modeling early language acquisition} {Baby
  {SRL}: Modeling early language acquisition}.{\BBCQ}
\newblock
\BIn{} A.~Clark\ \BBA {} K.~Toutanova\ (\BEDS), \APACrefbtitle {Proceedings of
  the {T}welfth {C}onference on {C}omputational {N}atural {L}anguage {L}earning
  ({CoNLL} 2008)} {Proceedings of the {T}welfth {C}onference on {C}omputational
  {N}atural {L}anguage {L}earning ({CoNLL} 2008)}\ (\BPGS\ 81--88).
\newblock
\APACaddressPublisher{}{Coling 2008 Organizing Committee}.
\newblock
\begin{APACrefURL} \url{https://aclanthology.org/W08-2111/} \end{APACrefURL}
\PrintBackRefs{\CurrentBib}

\bibitem [\protect \citeauthoryear {%
Conner%
, Gertner%
, Fisher%
\BCBL {}\ \BBA {} Roth%
}{%
Conner%
\ \protect \BOthers {.}}{%
{\protect \APACyear {2009}}%
}]{%
Conner:2009}
\APACinsertmetastar {%
Conner:2009}%
\begin{APACrefauthors}%
Conner, M.%
, Gertner, Y.%
, Fisher, C.%
\BCBL {}\ \BBA {} Roth, D.%
\end{APACrefauthors}%
\unskip\
\newblock
\APACrefYearMonthDay{2009}{}{}.
\newblock
{\BBOQ}\APACrefatitle {Minimally supervised model of early language
  acquisition} {Minimally supervised model of early language
  acquisition}.{\BBCQ}
\newblock
\BIn{} S.~Stevenson\ \BBA {} X.~Carreras\ (\BEDS), \APACrefbtitle
  {{P}roceedings of the {T}hirteenth {C}onference on {C}omputational {N}atural
  {L}anguage {L}earning ({CoNLL} 2009)} {{P}roceedings of the {T}hirteenth
  {C}onference on {C}omputational {N}atural {L}anguage {L}earning ({CoNLL}
  2009)}\ (\BPGS\ 84--92).
\newblock
\APACaddressPublisher{}{Association for Computational Linguistics (ACL)}.
\newblock
\begin{APACrefURL} \url{https://aclanthology.org/W09-1112} \end{APACrefURL}
\PrintBackRefs{\CurrentBib}

\bibitem [\protect \citeauthoryear {%
Cremers%
}{%
Cremers%
}{%
{\protect \APACyear {2002}}%
}]{%
Cremers:2002}
\APACinsertmetastar {%
Cremers:2002}%
\begin{APACrefauthors}%
Cremers, C.%
\end{APACrefauthors}%
\unskip\
\newblock
\APACrefYearMonthDay{2002}{}{}.
\newblock
{\BBOQ}\APACrefatitle {(’n) {B}etekenis berekend} {(’n) {B}etekenis
  berekend}.{\BBCQ}
\newblock
\APACjournalVolNumPages{Nederlandse Taalkunde}{7}{4}{375--395}.
\newblock
\begin{APACrefURL}
  \url{https://journal-archive.aup.nl/nederlandse-taalkunde/taalk\_2002\_nr4.pdf}
  \end{APACrefURL}
\PrintBackRefs{\CurrentBib}

\bibitem [\protect \citeauthoryear {%
Cremers%
\ \BBA {} Hijzelendoorn%
}{%
Cremers%
\ \BBA {} Hijzelendoorn%
}{%
{\protect \APACyear {1995/2025}}%
}]{%
Cremers:1995}
\APACinsertmetastar {%
Cremers:1995}%
\begin{APACrefauthors}%
Cremers, C.%
\BCBT {}\ \BBA {} Hijzelendoorn, P\BPBI M.%
\end{APACrefauthors}%
\unskip\
\newblock
\APACrefYearMonthDay{1995/2025}{}{}.
\newblock
\APACrefbtitle {Delilah. {Computer} software.} {Delilah. {Computer} software.}
\newblock
\APACaddressPublisher{}{Department of General Linguistics/LUCL, Leiden
  University}.
\PrintBackRefs{\CurrentBib}

\bibitem [\protect \citeauthoryear {%
Cremers%
\ \BBA {} Hijzelendoorn%
}{%
Cremers%
\ \BBA {} Hijzelendoorn%
}{%
{\protect \APACyear {2025}}%
}]{%
Cremers:2025}
\APACinsertmetastar {%
Cremers:2025}%
\begin{APACrefauthors}%
Cremers, C.%
\BCBT {}\ \BBA {} Hijzelendoorn, P\BPBI M.%
\end{APACrefauthors}%
\unskip\
\newblock
\APACrefYearMonthDay{2025}{}{}.
\newblock
\APACrefbtitle {{Delilah does Dutch. Online} demo version.} {{Delilah does
  Dutch. Online} demo version.}
\newblock
\begin{APACrefURL} \url{https://delilah.universiteitleiden.nl/indexen.html}
  \end{APACrefURL}
\newblock
\APACrefnote{Accessed on 15 January 2025}
\PrintBackRefs{\CurrentBib}

\bibitem [\protect \citeauthoryear {%
Cremers%
, Hijzelendoorn%
\BCBL {}\ \BBA {} Reckman%
}{%
Cremers%
\ \protect \BOthers {.}}{%
{\protect \APACyear {2014}}%
}]{%
Cremers:2014}
\APACinsertmetastar {%
Cremers:2014}%
\begin{APACrefauthors}%
Cremers, C.%
, Hijzelendoorn, P\BPBI M.%
\BCBL {}\ \BBA {} Reckman, H\BPBI G\BPBI B.%
\end{APACrefauthors}%
\unskip\
\newblock
\APACrefYear{2014}.
\newblock
\APACrefbtitle {Meaning versus grammar} {Meaning versus grammar}.
\newblock
\APACaddressPublisher{Leiden}{Leiden University Press}.
\newblock
\begin{APACrefDOI} \doi{https://doi.org/10.1515/9789400601833} \end{APACrefDOI}
\PrintBackRefs{\CurrentBib}

\bibitem [\protect \citeauthoryear {%
Cremers%
\ \BBA {} Reckman%
}{%
Cremers%
\ \BBA {} Reckman%
}{%
{\protect \APACyear {2008}}%
}]{%
Cremers:2008}
\APACinsertmetastar {%
Cremers:2008}%
\begin{APACrefauthors}%
Cremers, C.%
\BCBT {}\ \BBA {} Reckman, H\BPBI G\BPBI B.%
\end{APACrefauthors}%
\unskip\
\newblock
\APACrefYearMonthDay{2008}{}{}.
\newblock
{\BBOQ}\APACrefatitle {Exploiting logical forms} {Exploiting logical
  forms}.{\BBCQ}
\newblock
\BIn{} S.~Verberne, H.~van Halteren\BCBL {}\ \BBA {} P\BHBI A\BPBI J\BPBI
  M.~Coppen\ (\BEDS), \APACrefbtitle {{C}omputational {L}inguistics in the
  {N}etherlands 2007 ({CLIN} 18)} {{C}omputational {L}inguistics in the
  {N}etherlands 2007 ({CLIN} 18)}\ (\BVOL~11, \BPGS\ 5--20).
\newblock
\APACaddressPublisher{}{LOT}.
\newblock
\begin{APACrefURL}
  \url{https://www.clinjournal.org/CLIN\_proceedings/XVIII/cremers.pdf}
  \end{APACrefURL}
\PrintBackRefs{\CurrentBib}

\bibitem [\protect \citeauthoryear {%
Daelemans%
\ \BBA {} {\noopsort{Bosch}}{van den Bosch}%
}{%
Daelemans%
\ \BBA {} {\noopsort{Bosch}}{van den Bosch}%
}{%
{\protect \APACyear {2005}}%
}]{%
Daelemans:2005}
\APACinsertmetastar {%
Daelemans:2005}%
\begin{APACrefauthors}%
Daelemans, W.%
\BCBT {}\ \BBA {} {\noopsort{Bosch}}{van den Bosch}, A.%
\end{APACrefauthors}%
\unskip\
\newblock
\APACrefYear{2005}.
\newblock
\APACrefbtitle {Memory-based language processing} {Memory-based language
  processing}.
\newblock
\APACaddressPublisher{Cambridge}{Cambridge University Press}.
\newblock
\begin{APACrefDOI} \doi{https://doi.org/10.1017/CBO9780511486579}
  \end{APACrefDOI}
\PrintBackRefs{\CurrentBib}

\bibitem [\protect \citeauthoryear {%
Devlin%
, Chang%
, Lee%
\BCBL {}\ \BBA {} Toutanova%
}{%
Devlin%
\ \protect \BOthers {.}}{%
{\protect \APACyear {2019}}%
}]{%
Devlin:2019}
\APACinsertmetastar {%
Devlin:2019}%
\begin{APACrefauthors}%
Devlin, J.%
, Chang, M\BHBI W.%
, Lee, K.%
\BCBL {}\ \BBA {} Toutanova, K.%
\end{APACrefauthors}%
\unskip\
\newblock
\APACrefYearMonthDay{2019}{}{}.
\newblock
{\BBOQ}\APACrefatitle {{BERT}: {P}re-training of deep bidirectional
  transformers for language understanding} {{BERT}: {P}re-training of deep
  bidirectional transformers for language understanding}.{\BBCQ}
\newblock
\BIn{} J.~Burstein, C.~Doran\BCBL {}\ \BBA {} T.~Solorio\ (\BEDS),
  \APACrefbtitle {{N}orth {A}merican {C}hapter of the {A}ssociation for
  {C}omputational {L}inguistics: {H}uman {L}anguage {T}echnologies 2019
  ({NAACL-HLT})} {{N}orth {A}merican {C}hapter of the {A}ssociation for
  {C}omputational {L}inguistics: {H}uman {L}anguage {T}echnologies 2019
  ({NAACL-HLT})}\ (\BVOL\ 1 (Long and Short Papers), \BPGS\ 4171--4186).
\newblock
\APACaddressPublisher{}{Association for Computational Linguistics (ACL)}.
\newblock
\begin{APACrefDOI} \doi{https://doi.org/10.18653/v1/N19-1423} \end{APACrefDOI}
\PrintBackRefs{\CurrentBib}

\bibitem [\protect \citeauthoryear {%
Duda%
, Hart%
\BCBL {}\ \BBA {} Stork%
}{%
Duda%
\ \protect \BOthers {.}}{%
{\protect \APACyear {2001}}%
}]{%
Duda:2001}
\APACinsertmetastar {%
Duda:2001}%
\begin{APACrefauthors}%
Duda, R\BPBI O.%
, Hart, P\BPBI E.%
\BCBL {}\ \BBA {} Stork, D\BPBI G.%
\end{APACrefauthors}%
\unskip\
\newblock
\APACrefYear{2001}.
\newblock
\APACrefbtitle {Pattern classification} {Pattern classification}\
  (\PrintOrdinal{2nd}\ \BEd).
\newblock
\APACaddressPublisher{New York, NY}{Wiley}.
\PrintBackRefs{\CurrentBib}

\bibitem [\protect \citeauthoryear {%
Griffith%
\ \BBA {} Kalish%
}{%
Griffith%
\ \BBA {} Kalish%
}{%
{\protect \APACyear {2007}}%
}]{%
Griffith:2007}
\APACinsertmetastar {%
Griffith:2007}%
\begin{APACrefauthors}%
Griffith, T\BPBI L.%
\BCBT {}\ \BBA {} Kalish, M\BPBI L.%
\end{APACrefauthors}%
\unskip\
\newblock
\APACrefYearMonthDay{2007}{}{}.
\newblock
{\BBOQ}\APACrefatitle {Language evolution by iterated learning with {B}ayesian
  agents} {Language evolution by iterated learning with {B}ayesian
  agents}.{\BBCQ}
\newblock
\APACjournalVolNumPages{Cognitive Science}{31}{3}{441--480}.
\newblock
\begin{APACrefDOI} \doi{https://doi.org/10.1080/15326900701326576}
  \end{APACrefDOI}
\PrintBackRefs{\CurrentBib}

\bibitem [\protect \citeauthoryear {%
Gui%
\ \protect \BOthers {.}}{%
Gui%
\ \protect \BOthers {.}}{%
{\protect \APACyear {2024}}%
}]{%
Gui:2024}
\APACinsertmetastar {%
Gui:2024}%
\begin{APACrefauthors}%
Gui, J.%
, Chen, T.%
, Zhang, J.%
, Cao, Q.%
, Sun, Z.%
, Luo, H.%
\BCBL {}\ \BBA {} Tao, D.%
\end{APACrefauthors}%
\unskip\
\newblock
\APACrefYearMonthDay{2024}{}{}.
\newblock
{\BBOQ}\APACrefatitle {A survey on self-supervised learning: {A}lgorithms,
  applications, and future trends} {A survey on self-supervised learning:
  {A}lgorithms, applications, and future trends}.{\BBCQ}
\newblock
\APACjournalVolNumPages{{IEEE} Transactions on Pattern Analysis and Machine
  Intelligence}{}{}{1--20}.
\newblock
\begin{APACrefDOI} \doi{https://doi.org/10.1109/TPAMI.2024.3415112}
  \end{APACrefDOI}
\PrintBackRefs{\CurrentBib}

\bibitem [\protect \citeauthoryear {%
Herv\'{e}%
}{%
Herv\'{e}%
}{%
{\protect \APACyear {2022}}%
}]{%
Herve:2022}
\APACinsertmetastar {%
Herve:2022}%
\begin{APACrefauthors}%
Herv\'{e}, M.%
\end{APACrefauthors}%
\unskip\
\newblock
\APACrefYearMonthDay{2022}{}{}.
\newblock
\APACrefbtitle {{RVA}ide{M}emoire: {T}esting and plotting procedures for
  biostatistics ({Version} 0.9-81-2). {R} package.} {{RVA}ide{M}emoire:
  {T}esting and plotting procedures for biostatistics ({Version} 0.9-81-2). {R}
  package.}
\newblock
\begin{APACrefDOI} \doi{https://doi.org/10.32614/CRAN.package.RVAideMemoire}
  \end{APACrefDOI}
\PrintBackRefs{\CurrentBib}

\bibitem [\protect \citeauthoryear {%
{\noopsort{Hoeve}}{ter Hoeve}%
, Kharitonov%
, Hupkes%
\BCBL {}\ \BBA {} Dupoux%
}{%
{\noopsort{Hoeve}}{ter Hoeve}%
\ \protect \BOthers {.}}{%
{\protect \APACyear {2022}}%
}]{%
Hoeve:2022}
\APACinsertmetastar {%
Hoeve:2022}%
\begin{APACrefauthors}%
{\noopsort{Hoeve}}{ter Hoeve}, M.%
, Kharitonov, E.%
, Hupkes, D.%
\BCBL {}\ \BBA {} Dupoux, E.%
\end{APACrefauthors}%
\unskip\
\newblock
\APACrefYearMonthDay{2022}{}{}.
\newblock
{\BBOQ}\APACrefatitle {Towards interactive language modeling} {Towards
  interactive language modeling}.{\BBCQ}
\newblock
\APACjournalVolNumPages{Computing Research Repository
  (CoRR)}{{arXiv:cs.CL/2112.11911v2}}{}{}.
\newblock
\begin{APACrefDOI} \doi{https://doi.org/10.48550/arXiv.2112.11911}
  \end{APACrefDOI}
\PrintBackRefs{\CurrentBib}

\bibitem [\protect \citeauthoryear {%
Lan%
\ \protect \BOthers {.}}{%
Lan%
\ \protect \BOthers {.}}{%
{\protect \APACyear {2020}}%
}]{%
Lan:2020}
\APACinsertmetastar {%
Lan:2020}%
\begin{APACrefauthors}%
Lan, Z.%
, Chen, M.%
, Goodman, S.%
, Gimpel, K.%
, Sharma, P.%
\BCBL {}\ \BBA {} Soricut, R.%
\end{APACrefauthors}%
\unskip\
\newblock
\APACrefYearMonthDay{2020}{}{}.
\newblock
{\BBOQ}\APACrefatitle {{ALBERT}: {A} lite {BERT} for self-supervised learning
  of language representations} {{ALBERT}: {A} lite {BERT} for self-supervised
  learning of language representations}.{\BBCQ}
\newblock
\BIn{} \APACrefbtitle {{E}ighth {I}nternational {C}onference on {L}earning
  {R}epresentations ({ICLR} 2020).} {{E}ighth {I}nternational {C}onference on
  {L}earning {R}epresentations ({ICLR} 2020).}
\newblock
\begin{APACrefURL} \url{https://openreview.net/pdf?id=H1eA7AEtvS}
  \end{APACrefURL}
\PrintBackRefs{\CurrentBib}

\bibitem [\protect \citeauthoryear {%
MacDonald%
\ \BBA {} Gardner%
}{%
MacDonald%
\ \BBA {} Gardner%
}{%
{\protect \APACyear {2000}}%
}]{%
MacDonald:2000}
\APACinsertmetastar {%
MacDonald:2000}%
\begin{APACrefauthors}%
MacDonald, P\BPBI L.%
\BCBT {}\ \BBA {} Gardner, R\BPBI C.%
\end{APACrefauthors}%
\unskip\
\newblock
\APACrefYearMonthDay{2000}{}{}.
\newblock
{\BBOQ}\APACrefatitle {Type {I} error rate comparisons of post hoc procedures
  for {"IxJ"} chi-square tables} {Type {I} error rate comparisons of post hoc
  procedures for {"IxJ"} chi-square tables}.{\BBCQ}
\newblock
\APACjournalVolNumPages{Educational and Psychological
  Measurement}{60}{5}{735--754}.
\newblock
\begin{APACrefDOI} \doi{http://doi.org/10.1177/00131640021970871}
  \end{APACrefDOI}
\PrintBackRefs{\CurrentBib}

\bibitem [\protect \citeauthoryear {%
M\"{a}chler%
, Rousseeuw%
, Struyf%
, Hubert%
\BCBL {}\ \BBA {} Hornik%
}{%
M\"{a}chler%
\ \protect \BOthers {.}}{%
{\protect \APACyear {2019}}%
}]{%
Maechler:2019}
\APACinsertmetastar {%
Maechler:2019}%
\begin{APACrefauthors}%
M\"{a}chler, M.%
, Rousseeuw, P\BPBI J.%
, Struyf, A.%
, Hubert, M.%
\BCBL {}\ \BBA {} Hornik, K.%
\end{APACrefauthors}%
\unskip\
\newblock
\APACrefYearMonthDay{2019}{}{}.
\newblock
\APACrefbtitle {Cluster: {C}luster analysis basics and extensions ({Version}
  2.1.0). {R} package.} {Cluster: {C}luster analysis basics and extensions
  ({Version} 2.1.0). {R} package.}
\newblock
\begin{APACrefDOI} \doi{https://doi.org/10.32614/CRAN.package.cluster}
  \end{APACrefDOI}
\PrintBackRefs{\CurrentBib}

\bibitem [\protect \citeauthoryear {%
MacWhinney%
}{%
MacWhinney%
}{%
{\protect \APACyear {2014}}%
}]{%
MacWhinney:2014}
\APACinsertmetastar {%
MacWhinney:2014}%
\begin{APACrefauthors}%
MacWhinney, B.%
\end{APACrefauthors}%
\unskip\
\newblock
\APACrefYear{2014}.
\newblock
\APACrefbtitle {The {CHILDES} project: {T}ools for analyzing talk} {The
  {CHILDES} project: {T}ools for analyzing talk}\ (\PrintOrdinal{3rd}\ \BEd).
\newblock
\APACaddressPublisher{New York, NY}{Routledge}.
\newblock
\begin{APACrefDOI} \doi{https://doi.org/10.4324/9781315805641} \end{APACrefDOI}
\PrintBackRefs{\CurrentBib}

\bibitem [\protect \citeauthoryear {%
Marr%
}{%
Marr%
}{%
{\protect \APACyear {1982/2010}}%
}]{%
Marr:1982}
\APACinsertmetastar {%
Marr:1982}%
\begin{APACrefauthors}%
Marr, D.%
\end{APACrefauthors}%
\unskip\
\newblock
\APACrefYear{1982/2010}.
\newblock
\APACrefbtitle {Vision: {A} computational investigation into the human
  representation and processing of visual information} {Vision: {A}
  computational investigation into the human representation and processing of
  visual information}.
\newblock
\APACaddressPublisher{Cambridge, MA}{MIT Press}.
\newblock
\begin{APACrefDOI}
  \doi{https://doi.org/10.7551/mitpress/9780262514620.001.0001}
  \end{APACrefDOI}
\PrintBackRefs{\CurrentBib}

\bibitem [\protect \citeauthoryear {%
Matusevych%
, Alishahi%
\BCBL {}\ \BBA {} Vogt%
}{%
Matusevych%
\ \protect \BOthers {.}}{%
{\protect \APACyear {2013}}%
}]{%
Matusevych:2013}
\APACinsertmetastar {%
Matusevych:2013}%
\begin{APACrefauthors}%
Matusevych, Y.%
, Alishahi, A.%
\BCBL {}\ \BBA {} Vogt, P\BPBI A.%
\end{APACrefauthors}%
\unskip\
\newblock
\APACrefYearMonthDay{2013}{}{}.
\newblock
{\BBOQ}\APACrefatitle {Automatic generation of naturalistic child-adult
  interaction data} {Automatic generation of naturalistic child-adult
  interaction data}.{\BBCQ}
\newblock
\APACjournalVolNumPages{Proceedings of the Annual Meeting of the Cognitive
  Science Society}{35}{}{2996--3001}.
\newblock
\begin{APACrefURL} \url{https://escholarship.org/uc/item/2t9414br}
  \end{APACrefURL}
\PrintBackRefs{\CurrentBib}

\bibitem [\protect \citeauthoryear {%
McCoy%
, Frank%
\BCBL {}\ \BBA {} Linzen%
}{%
McCoy%
\ \protect \BOthers {.}}{%
{\protect \APACyear {2020}}%
}]{%
McCoy:2020}
\APACinsertmetastar {%
McCoy:2020}%
\begin{APACrefauthors}%
McCoy, R\BPBI T.%
, Frank, R.%
\BCBL {}\ \BBA {} Linzen, T.%
\end{APACrefauthors}%
\unskip\
\newblock
\APACrefYearMonthDay{2020}{}{}.
\newblock
{\BBOQ}\APACrefatitle {Does syntax need to grow on trees? {S}ources of
  hierarchical inductive bias in sequence-to-sequence networks} {Does syntax
  need to grow on trees? {S}ources of hierarchical inductive bias in
  sequence-to-sequence networks}.{\BBCQ}
\newblock
\APACjournalVolNumPages{Transactions of the Association for Computational
  Linguistics}{8}{}{125--140}.
\newblock
\begin{APACrefDOI} \doi{https://doi.org/10.1162/tacl\_a\_00304}
  \end{APACrefDOI}
\PrintBackRefs{\CurrentBib}

\bibitem [\protect \citeauthoryear {%
Moortgat%
}{%
Moortgat%
}{%
{\protect \APACyear {1997}}%
}]{%
Moortgat:1997}
\APACinsertmetastar {%
Moortgat:1997}%
\begin{APACrefauthors}%
Moortgat, M\BPBI J.%
\end{APACrefauthors}%
\unskip\
\newblock
\APACrefYearMonthDay{1997}{}{}.
\newblock
{\BBOQ}\APACrefatitle {Categorial Type Logics} {Categorial type logics}.{\BBCQ}
\newblock
\BIn{} J\BPBI F\BPBI A\BPBI K.~{\noopsort{Benthem}}{van Benthem}\ \BBA {}
  A\BPBI G\BPBI B.~{\noopsort{Meulen}}{ter Meulen}\ (\BEDS), \APACrefbtitle
  {Handbook of Logic and Language} {Handbook of logic and language}\ (\BPGS\
  93--177).
\newblock
\APACaddressPublisher{Amsterdam}{Elsevier}.
\PrintBackRefs{\CurrentBib}

\bibitem [\protect \citeauthoryear {%
Orhan%
, Gupta%
\BCBL {}\ \BBA {} Lake%
}{%
Orhan%
\ \protect \BOthers {.}}{%
{\protect \APACyear {2020}}%
}]{%
Orhan:2020}
\APACinsertmetastar {%
Orhan:2020}%
\begin{APACrefauthors}%
Orhan, A\BPBI E.%
, Gupta, V\BPBI V.%
\BCBL {}\ \BBA {} Lake, B\BPBI M.%
\end{APACrefauthors}%
\unskip\
\newblock
\APACrefYearMonthDay{2020}{}{}.
\newblock
{\BBOQ}\APACrefatitle {Self-supervised learning through the eyes of a child}
  {Self-supervised learning through the eyes of a child}.{\BBCQ}
\newblock
\BIn{} H.~Larochelle, M.~Ranzato, R.~Hadsell, M\BHBI F.~Balcan\BCBL {}\ \BBA {}
  H\BHBI T.~Lin\ (\BEDS), \APACrefbtitle {{A}dvances in {N}eural {I}nformation
  {P}rocessing {S}ystems 33 ({NeurIPS} 2020)} {{A}dvances in {N}eural
  {I}nformation {P}rocessing {S}ystems 33 ({NeurIPS} 2020)}\ (\BVOL~33, \BPGS\
  9960--9971).
\newblock
\APACaddressPublisher{}{Curran Associates}.
\newblock
\begin{APACrefURL}
  \url{https://proceedings.neurips.cc/paper\_files/paper/2020/file/7183145a2a3e0ce2b68cd3735186b1d5-Paper.pdf}
  \end{APACrefURL}
\PrintBackRefs{\CurrentBib}

\bibitem [\protect \citeauthoryear {%
Pollard%
\ \BBA {} Sag%
}{%
Pollard%
\ \BBA {} Sag%
}{%
{\protect \APACyear {1987}}%
}]{%
Pollard:1987}
\APACinsertmetastar {%
Pollard:1987}%
\begin{APACrefauthors}%
Pollard, C\BPBI J.%
\BCBT {}\ \BBA {} Sag, I\BPBI A.%
\end{APACrefauthors}%
\unskip\
\newblock
\APACrefYear{1987}.
\newblock
\APACrefbtitle {Information-based syntax and semantics} {Information-based
  syntax and semantics}\ (\BVOL\ 1: Fundamentals).
\newblock
\APACaddressPublisher{Stanford, CA}{Center for the Study of Language and
  Information}.
\PrintBackRefs{\CurrentBib}

\bibitem [\protect \citeauthoryear {%
Pollard%
\ \BBA {} Sag%
}{%
Pollard%
\ \BBA {} Sag%
}{%
{\protect \APACyear {1994}}%
}]{%
Pollard:1994}
\APACinsertmetastar {%
Pollard:1994}%
\begin{APACrefauthors}%
Pollard, C\BPBI J.%
\BCBT {}\ \BBA {} Sag, I\BPBI A.%
\end{APACrefauthors}%
\unskip\
\newblock
\APACrefYear{1994}.
\newblock
\APACrefbtitle {Head-{D}riven {P}hrase {S}tructure {G}rammar} {Head-{D}riven
  {P}hrase {S}tructure {G}rammar}.
\newblock
\APACaddressPublisher{Stanford, CA and Chicago, IL}{Center for the Study of
  Language and Information and University of Chicago Press}.
\PrintBackRefs{\CurrentBib}

\bibitem [\protect \citeauthoryear {%
{R Core Team}%
}{%
{R Core Team}%
}{%
{\protect \APACyear {2020}}%
}]{%
RCoreTeam:2020}
\APACinsertmetastar {%
RCoreTeam:2020}%
\begin{APACrefauthors}%
{R Core Team}.%
\end{APACrefauthors}%
\unskip\
\newblock
\APACrefYearMonthDay{2020}{}{}.
\newblock
\APACrefbtitle {R: A language and environment for statistical computing
  ({Version} 3.6.3). {Computer} software.} {R: A language and environment for
  statistical computing ({Version} 3.6.3). {Computer} software.}
\newblock
\APACaddressPublisher{Vienna}{R Foundation for Statistical Computing}.
\newblock
\begin{APACrefURL} \url{http://www.R-project.org/} \end{APACrefURL}
\PrintBackRefs{\CurrentBib}

\bibitem [\protect \citeauthoryear {%
Reckman%
}{%
Reckman%
}{%
{\protect \APACyear {2009}}%
}]{%
Reckman:2009}
\APACinsertmetastar {%
Reckman:2009}%
\begin{APACrefauthors}%
Reckman, H\BPBI G\BPBI B.%
\end{APACrefauthors}%
\unskip\
\newblock
\APACrefYear{2009}.
\unskip\
\newblock
\APACrefbtitle {Flat but not shallow: {T}owards flatter representations in deep
  semantic parsing for precise and feasible inferencing} {Flat but not shallow:
  {T}owards flatter representations in deep semantic parsing for precise and
  feasible inferencing}\ \APACtypeAddressSchool {{PhD} thesis, {LOT}
  {D}issertation {S}eries 203}{}{Leiden University}.
\unskip\
\newblock
\APAChowpublished {Utrecht: {LOT}}.
\unskip\
\newblock
\begin{APACrefURL}
  \url{https://www.lotpublications.nl/flat-but-not-shallow-flat-but-not-shallow-towards-flatter-representations-in-deep-semantic-parsing-for-precise-and-feasible-inferencing}
  \end{APACrefURL}
\PrintBackRefs{\CurrentBib}

\bibitem [\protect \citeauthoryear {%
Rita%
, Chaabouni%
\BCBL {}\ \BBA {} Dupoux%
}{%
Rita%
\ \protect \BOthers {.}}{%
{\protect \APACyear {2020}}%
}]{%
Rita:2020}
\APACinsertmetastar {%
Rita:2020}%
\begin{APACrefauthors}%
Rita, M.%
, Chaabouni, R.%
\BCBL {}\ \BBA {} Dupoux, E.%
\end{APACrefauthors}%
\unskip\
\newblock
\APACrefYearMonthDay{2020}{}{}.
\newblock
{\BBOQ}\APACrefatitle {{“LazImpa”}: {L}azy and {I}mpatient neural agents
  learn to communicate efficiently} {{“LazImpa”}: {L}azy and {I}mpatient
  neural agents learn to communicate efficiently}.{\BBCQ}
\newblock
\BIn{} R.~Fern\'{a}ndez\ \BBA {} T.~Linzen\ (\BEDS), \APACrefbtitle
  {{P}roceedings of the 24th {C}onference on {C}omputational {N}atural
  {L}anguage {L}earning ({CoNLL} 2020)} {{P}roceedings of the 24th {C}onference
  on {C}omputational {N}atural {L}anguage {L}earning ({CoNLL} 2020)}\ (\BPGS\
  335--343).
\newblock
\APACaddressPublisher{}{Association for Computational Linguistics (ACL)}.
\newblock
\begin{APACrefDOI} \doi{https://doi.org/10.18653/v1/2020.conll-1.26}
  \end{APACrefDOI}
\PrintBackRefs{\CurrentBib}

\bibitem [\protect \citeauthoryear {%
Sch\"{u}tze%
}{%
Sch\"{u}tze%
}{%
{\protect \APACyear {1995}}%
}]{%
Schutze:1995}
\APACinsertmetastar {%
Schutze:1995}%
\begin{APACrefauthors}%
Sch\"{u}tze, H.%
\end{APACrefauthors}%
\unskip\
\newblock
\APACrefYearMonthDay{1995}{}{}.
\newblock
{\BBOQ}\APACrefatitle {Distributional part-of-speech tagging} {Distributional
  part-of-speech tagging}.{\BBCQ}
\newblock
\BIn{} S\BPBI P.~Abney\ \BBA {} E\BPBI W.~Hinrichs\ (\BEDS), \APACrefbtitle
  {Proceedings of the {S}eventh {C}onference of the {E}uropean {C}hapter of the
  {A}ssociation for {C}omputational {L}inguistics ({EACL} '95)} {Proceedings of
  the {S}eventh {C}onference of the {E}uropean {C}hapter of the {A}ssociation
  for {C}omputational {L}inguistics ({EACL} '95)}\ (\BPGS\ 141--148).
\newblock
\APACaddressPublisher{San Francisco, CA}{Morgan Kaufman}.
\newblock
\begin{APACrefDOI} \doi{https://doi.org/10.3115/976973.976994} \end{APACrefDOI}
\PrintBackRefs{\CurrentBib}

\bibitem [\protect \citeauthoryear {%
Shakouri%
, Cremers%
\BCBL {}\ \BBA {} Schiller%
}{%
Shakouri%
\ \protect \BOthers {.}}{%
{\protect \APACyear {2025}}%
}]{%
Shakouri:2025}
\APACinsertmetastar {%
Shakouri:2025}%
\begin{APACrefauthors}%
Shakouri, D\BPBI P.%
, Cremers, C.%
\BCBL {}\ \BBA {} Schiller, N\BPBI O.%
\end{APACrefauthors}%
\unskip\
\newblock
\APACrefYearMonthDay{2025}{}{}.
\newblock
{\BBOQ}\APACrefatitle {A knowledge-based language model: {D}educing grammatical
  knowledge in a multi-agent language acquisition simulation} {A
  knowledge-based language model: {D}educing grammatical knowledge in a
  multi-agent language acquisition simulation}.{\BBCQ}
\newblock
\APACjournalVolNumPages{Computational Linguistics in the Netherlands
  Journal}{14}{}{167--189}.
\newblock
\begin{APACrefURL} \url{https://www.clinjournal.org/clinj/article/view/193}
  \end{APACrefURL}
\PrintBackRefs{\CurrentBib}

\bibitem [\protect \citeauthoryear {%
Steedman%
}{%
Steedman%
}{%
{\protect \APACyear {1996}}%
}]{%
Steedman:1996}
\APACinsertmetastar {%
Steedman:1996}%
\begin{APACrefauthors}%
Steedman, M.%
\end{APACrefauthors}%
\unskip\
\newblock
\APACrefYear{1996}.
\newblock
\APACrefbtitle {Surface structure and interpretation} {Surface structure and
  interpretation}.
\newblock
\APACaddressPublisher{Cambridge, MA}{MIT Press}.
\PrintBackRefs{\CurrentBib}

\bibitem [\protect \citeauthoryear {%
Steels%
}{%
Steels%
}{%
{\protect \APACyear {2000}}%
}]{%
Steels:2000}
\APACinsertmetastar {%
Steels:2000}%
\begin{APACrefauthors}%
Steels, L.%
\end{APACrefauthors}%
\unskip\
\newblock
\APACrefYearMonthDay{2000}{}{}.
\newblock
{\BBOQ}\APACrefatitle {The emergence of grammar in communicating autonomous
  robotic agents} {The emergence of grammar in communicating autonomous robotic
  agents}.{\BBCQ}
\newblock
\BIn{} W.~Horn\ (\BED), \APACrefbtitle {Proceedings of the 14th {E}uropean
  {C}onference on {A}rtificial {I}ntelligence 2000 ({ECAI 2000})} {Proceedings
  of the 14th {E}uropean {C}onference on {A}rtificial {I}ntelligence 2000
  ({ECAI 2000})}\ (\BPGS\ 764--769).
\newblock
\APACaddressPublisher{Amsterdam}{IOS}.
\newblock
\begin{APACrefURL} \url{https://frontiersinai.com/ecai/ecai2000/pdf/p0764.pdf}
  \end{APACrefURL}
\PrintBackRefs{\CurrentBib}

\bibitem [\protect \citeauthoryear {%
Steels%
}{%
Steels%
}{%
{\protect \APACyear {2001}}%
}]{%
Steels:2001}
\APACinsertmetastar {%
Steels:2001}%
\begin{APACrefauthors}%
Steels, L.%
\end{APACrefauthors}%
\unskip\
\newblock
\APACrefYearMonthDay{2001}{}{}.
\newblock
{\BBOQ}\APACrefatitle {Language games for autonomous robots} {Language games
  for autonomous robots}.{\BBCQ}
\newblock
\APACjournalVolNumPages{{IEEE} Intelligent Systems}{16}{5}{16--22}.
\newblock
\begin{APACrefDOI} \doi{https://doi.org/10.1109/5254.956077} \end{APACrefDOI}
\PrintBackRefs{\CurrentBib}

\bibitem [\protect \citeauthoryear {%
Steels%
}{%
Steels%
}{%
{\protect \APACyear {2015}}%
}]{%
Steels:2015}
\APACinsertmetastar {%
Steels:2015}%
\begin{APACrefauthors}%
Steels, L.%
\end{APACrefauthors}%
\unskip\
\newblock
\APACrefYear{2015}.
\newblock
\APACrefbtitle {The {T}alking {H}eads experiment: {O}rigins of words and
  meanings} {The {T}alking {H}eads experiment: {O}rigins of words and
  meanings}.
\newblock
\APACaddressPublisher{Berlin}{Language Science}.
\newblock
\begin{APACrefURL}
  \url{https://doi.org/10.17169/FUDOCS\_document\_000000022455}
  \end{APACrefURL}
\PrintBackRefs{\CurrentBib}

\bibitem [\protect \citeauthoryear {%
Steels%
\ \BBA {} Beuls%
}{%
Steels%
\ \BBA {} Beuls%
}{%
{\protect \APACyear {2017}}%
}]{%
Steels:2017}
\APACinsertmetastar {%
Steels:2017}%
\begin{APACrefauthors}%
Steels, L.%
\BCBT {}\ \BBA {} Beuls, K.%
\end{APACrefauthors}%
\unskip\
\newblock
\APACrefYearMonthDay{2017}{}{}.
\newblock
{\BBOQ}\APACrefatitle {How to explain the origins of complexity in language: A
  case study for agreement systems} {How to explain the origins of complexity
  in language: A case study for agreement systems}.{\BBCQ}
\newblock
\BIn{} S\BPBI S.~Mufwene, C.~Coup\'{e}\BCBL {}\ \BBA {} F.~Pellegrino\ (\BEDS),
  \APACrefbtitle {Complexity in language: {D}evelopmental and evolutionary
  perspectives} {Complexity in language: {D}evelopmental and evolutionary
  perspectives}\ (\BPGS\ 30--47).
\newblock
\APACaddressPublisher{Cambridge}{Cambridge University Press}.
\newblock
\begin{APACrefDOI} \doi{https://doi.org/10.1017/9781107294264.002}
  \end{APACrefDOI}
\PrintBackRefs{\CurrentBib}

\bibitem [\protect \citeauthoryear {%
Steels%
, {\noopsort{Eecke}}{van Eecke}%
\BCBL {}\ \BBA {} Beuls%
}{%
Steels%
\ \protect \BOthers {.}}{%
{\protect \APACyear {2018}}%
}]{%
Steels:2018}
\APACinsertmetastar {%
Steels:2018}%
\begin{APACrefauthors}%
Steels, L.%
, {\noopsort{Eecke}}{van Eecke}, P.%
\BCBL {}\ \BBA {} Beuls, K.%
\end{APACrefauthors}%
\unskip\
\newblock
\APACrefYearMonthDay{2018}{}{}.
\newblock
{\BBOQ}\APACrefatitle {Usage-based learning of grammatical categories}
  {Usage-based learning of grammatical categories}.{\BBCQ}
\newblock
\BIn{} M.~Atzmueller\ \BBA {} W.~Duivesteijn\ (\BEDS), \APACrefbtitle
  {Preproceedings of the 30th {B}enelux {C}onference on {A}rtificial
  {I}ntelligence ({BNAIC} 2018)} {Preproceedings of the 30th {B}enelux
  {C}onference on {A}rtificial {I}ntelligence ({BNAIC} 2018)}\ (\BPGS\
  253--264).
\newblock
\APACaddressPublisher{’s-Hertogenbosch}{Jheronimus Bosch Academy of Data
  Science (JADS)}.
\PrintBackRefs{\CurrentBib}

\bibitem [\protect \citeauthoryear {%
Steels%
\ \BBA {} Kaplan%
}{%
Steels%
\ \BBA {} Kaplan%
}{%
{\protect \APACyear {2001}}%
}]{%
SteelsKaplan:2001}
\APACinsertmetastar {%
SteelsKaplan:2001}%
\begin{APACrefauthors}%
Steels, L.%
\BCBT {}\ \BBA {} Kaplan, F.%
\end{APACrefauthors}%
\unskip\
\newblock
\APACrefYearMonthDay{2001}{}{}.
\newblock
{\BBOQ}\APACrefatitle {{AIBO}’s first words: The social learning of language
  and meaning} {{AIBO}’s first words: The social learning of language and
  meaning}.{\BBCQ}
\newblock
\APACjournalVolNumPages{Evolution of Communication}{4}{1}{3--32}.
\newblock
\begin{APACrefDOI} \doi{https://doi.org/10.1075/eoc.4.1.03ste} \end{APACrefDOI}
\PrintBackRefs{\CurrentBib}

\bibitem [\protect \citeauthoryear {%
Steels%
\ \BBA {} Loetzch%
}{%
Steels%
\ \BBA {} Loetzch%
}{%
{\protect \APACyear {2012}}%
}]{%
Steels:2012}
\APACinsertmetastar {%
Steels:2012}%
\begin{APACrefauthors}%
Steels, L.%
\BCBT {}\ \BBA {} Loetzch, M.%
\end{APACrefauthors}%
\unskip\
\newblock
\APACrefYearMonthDay{2012}{}{}.
\newblock
{\BBOQ}\APACrefatitle {The grounded naming game} {The grounded naming
  game}.{\BBCQ}
\newblock
\BIn{} L.~Steels\ (\BED), \APACrefbtitle {Experiments in cultural language
  evolution} {Experiments in cultural language evolution}\ (\BVOL~3, \BPGS\
  41--60).
\newblock
\APACaddressPublisher{Amsterdam}{John Benjamins}.
\newblock
\begin{APACrefDOI} \doi{http://doi.org/10.1075\%2Fais.3.04ste} \end{APACrefDOI}
\PrintBackRefs{\CurrentBib}

\bibitem [\protect \citeauthoryear {%
{\noopsort{Trijp}}{van Trijp}%
}{%
{\noopsort{Trijp}}{van Trijp}%
}{%
{\protect \APACyear {2016}}%
}]{%
Trijp:2016}
\APACinsertmetastar {%
Trijp:2016}%
\begin{APACrefauthors}%
{\noopsort{Trijp}}{van Trijp}, R.%
\end{APACrefauthors}%
\unskip\
\newblock
\APACrefYear{2016}.
\newblock
\APACrefbtitle {The evolution of case grammar} {The evolution of case grammar}.
\newblock
\APACaddressPublisher{Berlin}{Language Science}.
\newblock
\begin{APACrefDOI} \doi{https://doi.org/10.17169/langsci.b52.182}
  \end{APACrefDOI}
\PrintBackRefs{\CurrentBib}

\bibitem [\protect \citeauthoryear {%
Urbanek%
}{%
Urbanek%
}{%
{\protect \APACyear {2009}}%
}]{%
Urbanek:2009}
\APACinsertmetastar {%
Urbanek:2009}%
\begin{APACrefauthors}%
Urbanek, S.%
\end{APACrefauthors}%
\unskip\
\newblock
\APACrefYearMonthDay{2009}{}{}.
\newblock
{\BBOQ}\APACrefatitle {How to talk to strangers: {W}ays to leverage
  connectivity between {R, Java and Objective C}} {How to talk to strangers:
  {W}ays to leverage connectivity between {R, Java and Objective C}}.{\BBCQ}
\newblock
\APACjournalVolNumPages{Computational Statistics}{24}{2}{303--311}.
\newblock
\begin{APACrefDOI} \doi{https://doi.org/10.1007/s00180-008-0132-x}
  \end{APACrefDOI}
\PrintBackRefs{\CurrentBib}

\bibitem [\protect \citeauthoryear {%
Urbanek%
}{%
Urbanek%
}{%
{\protect \APACyear {2012}}%
}]{%
Urbanek:2012}
\APACinsertmetastar {%
Urbanek:2012}%
\begin{APACrefauthors}%
Urbanek, S.%
\end{APACrefauthors}%
\unskip\
\newblock
\APACrefYearMonthDay{2012}{}{}.
\newblock
\APACrefbtitle {{JavaGD}: {J}ava graphics device ({Version} 0.6-1). {R}
  package.} {{JavaGD}: {J}ava graphics device ({Version} 0.6-1). {R} package.}
\newblock
\begin{APACrefDOI} \doi{https://doi.org/10.32614/CRAN.package.JavaGD}
  \end{APACrefDOI}
\PrintBackRefs{\CurrentBib}

\bibitem [\protect \citeauthoryear {%
Urbanek%
}{%
Urbanek%
}{%
{\protect \APACyear {2017}}%
}]{%
Urbanek:2017}
\APACinsertmetastar {%
Urbanek:2017}%
\begin{APACrefauthors}%
Urbanek, S.%
\end{APACrefauthors}%
\unskip\
\newblock
\APACrefYearMonthDay{2017}{}{}.
\newblock
\APACrefbtitle {{RJava}: {L}ow-level {R} to {Java} interface ({Version} 0.9-9).
  {R} package.} {{RJava}: {L}ow-level {R} to {Java} interface ({Version}
  0.9-9). {R} package.}
\newblock
\begin{APACrefDOI} \doi{https://doi.org/10.32614/CRAN.package.rJava}
  \end{APACrefDOI}
\PrintBackRefs{\CurrentBib}

\bibitem [\protect \citeauthoryear {%
Wickham%
}{%
Wickham%
}{%
{\protect \APACyear {2011}}%
}]{%
Wickham:2011}
\APACinsertmetastar {%
Wickham:2011}%
\begin{APACrefauthors}%
Wickham, H.%
\end{APACrefauthors}%
\unskip\
\newblock
\APACrefYearMonthDay{2011}{}{}.
\newblock
{\BBOQ}\APACrefatitle {The split-apply-combine strategy for data analysis} {The
  split-apply-combine strategy for data analysis}.{\BBCQ}
\newblock
\APACjournalVolNumPages{Journal of Statistical Software}{40}{1}{1--29}.
\newblock
\begin{APACrefDOI} \doi{https://doi.org/10.18637/jss.v040.i01} \end{APACrefDOI}
\PrintBackRefs{\CurrentBib}

\bibitem [\protect \citeauthoryear {%
Wickham%
}{%
Wickham%
}{%
{\protect \APACyear {2020}}%
}]{%
Wickham:2020a}
\APACinsertmetastar {%
Wickham:2020a}%
\begin{APACrefauthors}%
Wickham, H.%
\end{APACrefauthors}%
\unskip\
\newblock
\APACrefYearMonthDay{2020}{}{}.
\newblock
\APACrefbtitle {Plyr: {T}ools for splitting, applying and combining data
  ({Version} 1.8.6). {R} package.} {Plyr: {T}ools for splitting, applying and
  combining data ({Version} 1.8.6). {R} package.}
\newblock
\begin{APACrefDOI} \doi{https://doi.org/10.32614/CRAN.package.plyr}
  \end{APACrefDOI}
\PrintBackRefs{\CurrentBib}

\bibitem [\protect \citeauthoryear {%
Wickham%
\ \BBA {} Grolemund%
}{%
Wickham%
\ \BBA {} Grolemund%
}{%
{\protect \APACyear {2016}}%
}]{%
Wickham:2016}
\APACinsertmetastar {%
Wickham:2016}%
\begin{APACrefauthors}%
Wickham, H.%
\BCBT {}\ \BBA {} Grolemund, G.%
\end{APACrefauthors}%
\unskip\
\newblock
\APACrefYear{2016}.
\newblock
\APACrefbtitle {R for data science: {I}mport, tidy, transform, visualize, and
  model data} {R for data science: {I}mport, tidy, transform, visualize, and
  model data}\ (\PrintOrdinal{1st}\ \BEd).
\newblock
\APACaddressPublisher{Sebastopol, CA}{O'Reilly Media}.
\PrintBackRefs{\CurrentBib}

\bibitem [\protect \citeauthoryear {%
Wickham%
, Vaughan%
\BCBL {}\ \BBA {} Girlich%
}{%
Wickham%
\ \protect \BOthers {.}}{%
{\protect \APACyear {2020}}%
}]{%
Wickham:2020b}
\APACinsertmetastar {%
Wickham:2020b}%
\begin{APACrefauthors}%
Wickham, H.%
, Vaughan, D.%
\BCBL {}\ \BBA {} Girlich, M.%
\end{APACrefauthors}%
\unskip\
\newblock
\APACrefYearMonthDay{2020}{}{}.
\newblock
\APACrefbtitle {Tidyr: {T}idy messy data ({Version} 1.1.1). {R} package.}
  {Tidyr: {T}idy messy data ({Version} 1.1.1). {R} package.}
\newblock
\begin{APACrefDOI} \doi{https://doi.org/10.32614/CRAN.package.tidyr}
  \end{APACrefDOI}
\PrintBackRefs{\CurrentBib}

\bibitem [\protect \citeauthoryear {%
Yarowsky%
}{%
Yarowsky%
}{%
{\protect \APACyear {1995}}%
}]{%
Yarowsky:1995}
\APACinsertmetastar {%
Yarowsky:1995}%
\begin{APACrefauthors}%
Yarowsky, D.%
\end{APACrefauthors}%
\unskip\
\newblock
\APACrefYearMonthDay{1995}{}{}.
\newblock
{\BBOQ}\APACrefatitle {Unsupervised word sense disambiguation rivaling
  supervised methods} {Unsupervised word sense disambiguation rivaling
  supervised methods}.{\BBCQ}
\newblock
\BIn{} \APACrefbtitle {{P}roceedings of the 33rd {A}nnual {M}eeting of the
  {A}ssociation for {C}omputational {L}inguistics} {{P}roceedings of the 33rd
  {A}nnual {M}eeting of the {A}ssociation for {C}omputational {L}inguistics}\
  (\BPGS\ 189--196).
\newblock
\APACaddressPublisher{}{Association for Computational Linguistics (ACL)}.
\newblock
\begin{APACrefDOI} \doi{https://doi.org/10.3115/981658.981684} \end{APACrefDOI}
\PrintBackRefs{\CurrentBib}

\end{thebibliography}

\appendix
\section*{Appendix A. Glossary of Grammatical Terms and Abbreviations}
\label{appendix:A}
\setcounter{table}{0}
\renewcommand{\thetable}{A.\arabic{table}}
\begin{center}
\begin{tabular}{ l l }
\hline
\textbf{Term / Abbreviation} & \textbf{Description} \\ \hline
\textsc{1} & first person \\
\textsc{2} & second person \\
\textsc{3} & third person \\
\textsc{adj} & adjective \\
\textsc{archaic} & archaic \\
\textsc{clit} & clitic \\
\textsc{comp} & complementizer \\
\textsc{det} & determiner \\
\textsc{fem} & feminine \\
\textsc{formal} & formal \\
\textsc{gen} & genitive \\
\textsc{inf} & infinitive \\
\textsc{inform} & informal \\
\textsc{masc} & masculine \\
\textsc{n} & noun \\
\textsc{neut} & neuter \\
\textsc{part} & participle \\
\textsc{past} & past \\
\textsc{pl} & plural \\
\textsc{prep} & preposition \\
\textsc{pres} & present \\
\textsc{sg} & singular \\
\textsc{v} & verb \\ \hline
\end{tabular}
\captionof{table}{Glossary of grammatical abbreviations}
\label{tab:app_gloss}
\end{center}
\end{document}